\documentclass{article}

\usepackage{arxiv}

\usepackage{amsmath,amsfonts,bm}

\def\eqref#1{equation~\ref{#1}}

\def\1{\bm{1}}

\def\vzero{{\bm{0}}}

\def\vs{{\bm{s}}}

\def\vx{{\bm{x}}}

\def\mX{{\bm{X}}}

\DeclareMathAlphabet{\mathsfit}{\encodingdefault}{\sfdefault}{m}{sl}
\SetMathAlphabet{\mathsfit}{bold}{\encodingdefault}{\sfdefault}{bx}{n}

\usepackage{hyperref}
\hypersetup{
    colorlinks=true,
    citecolor=blue
}

\usepackage{url}
\usepackage{booktabs}
\usepackage{graphicx}
\usepackage{algorithm} 
\usepackage{algpseudocode}
\usepackage{varwidth}
\usepackage{multicol}
\usepackage{multirow}
\usepackage{caption}
\usepackage{subcaption}
\usepackage{wrapfig}

\usepackage[utf8]{inputenc} 
\usepackage[T1]{fontenc}    
\usepackage{hyperref}       
\usepackage{url}            
\usepackage{booktabs}       
\usepackage{amsfonts}       
\usepackage{nicefrac}       
\usepackage{microtype}      
\usepackage{lipsum}
\usepackage{amssymb}
\usepackage{pifont}
\usepackage{xcolor}
\usepackage{longtable}

\newcommand{\fref}[1]{Figure~\ref{#1}}

\newcommand{\tref}[1]{Table~\ref{#1}}

\newcommand{\xmark}{{\color[HTML]{CB0000} \text{\ding{55}}}}

\title{MIMIC-IF: Interpretability and Fairness Evaluation of Deep Learning Models on MIMIC-IV Dataset}

\author{
  Chuizheng Meng\thanks{Authors contributed equally.},\quad Loc Trinh\footnotemark[1],\quad Nan Xu\footnotemark[1],\quad Yan Liu \\
  Department of Computer Science\\
  University of Southern California\\
  Los Angeles, CA 90089 \\
  \texttt{\{chuizhem, loctrinh, nanx, yanliu.cs\}@usc.edu} \\
}

\begin{document}
\maketitle

\begin{abstract}

    The recent release of large-scale healthcare datasets has greatly propelled the research of data-driven deep learning models for healthcare applications. However, due to the nature of such deep black-boxed models, concerns about interpretability, fairness, and biases in healthcare scenarios where human lives are at stake call for a careful and thorough examinations of both datasets and models. In this work, we focus on MIMIC-IV (Medical Information Mart for Intensive Care, version IV), the largest publicly available healthcare dataset, and conduct comprehensive analyses of dataset representation bias as well as interpretability and prediction fairness of deep learning models for in-hospital mortality prediction.
 In terms of interpretabilty, we observe that (1) the best performing interpretability method successfully identifies critical features for mortality prediction on various prediction models; (2) demographic features are important for prediction.
    In terms of fairness, we observe that (1) there exists disparate treatment in prescribing mechanical ventilation among patient groups across ethnicity, gender and age; (2) all of the studied mortality predictors are generally fair while the IMV-LSTM (Interpretable Multi-Variable Long Short-Term Memory) model provides the most accurate and unbiased predictions across all protected groups. We further draw concrete connections between interpretability methods and fairness metrics by showing how feature importance from interpretability methods can be beneficial in quantifying potential disparities in mortality predictors.
\end{abstract}

\section{Introduction}
With the release of large scale healthcare datasets, research of data-driven deep learning methods for healthcare applications demonstrates their superior performance over traditional methods on various tasks, including mortality prediction, length-of-stay prediction, phenotyping classification and intervention prediction~\cite{PURUSHOTHAM2018112,harutyunyan2019multitask,wang2020mimic}.
However, deep learning models have been treated as black-box universal function approximators, where prediction explanations are no longer available as their traditional counterparts, e.g., Logistic Regression and Random Forests. Lack of interpretability hinders the wide application of deep learning models in critical domains like healthcare.
In addition, due to bias in datasets or models, decisions made by machine learning algorithms are prone to be unfair, where an individual or a group is favored compared with the others owing to their inherent traits. As a result, more and more concerns about interpretability, fairness and biases have been raised recently in the healthcare domain where human lives are at stake~\cite{chen2018my}.
These concerns call for careful and thorough analyses of both datasets and algorithms.


In this work, we focus on the latest version (version IV~\cite{johnson2020mimiciv}) of a widely used large scale healthcare dataset MIMIC~\cite{goldberger2000physionet}, and conduct comprehensive analyses of model interpretability, dataset bias, algorithmic fairness, and the interaction between interpretability and fairness. 
\paragraph{Interpretability evaluation.} First, we evaluate the performance of common interpretability methods for feature importance estimation on multiple deep learning models trained for the mortality prediction task. Due to the complexity of dynamics in electronic health record data, there is no access to the ground truth of feature importance. Therefore, we utilize ROAR (remove and retrain)~\cite{hooker2019benchmark} to quantitatively evaluate different feature importance estimations. On all models considered, the ArchDetect~\cite{tsang2020does} outperforms other interpretation methods in feature importance estimation.Then we qualitatively analyze the feature importance estimation results given by ArchDetect, and verify its effectiveness based on the observations that it successfully identifies critical features for mortality prediction. We also find that demographic features are important for prediction, which leads to our following analyses of dataset bias and algorithmic fairness.
\paragraph{Dataset bias and algorithmic fairness.} 
We adopt the following commonly used demographic features as protected attributes: 1) \emph{ethnicity}, 2) \emph{gender}, 3) \emph{marital status}, 4) \emph{age}, and 5) \emph{insurance type}. For dataset bias, we analyze the average adoption and duration of five types of ventilation treatment on patients from different groups. There exists treatment disparity among patient groups split by different protected attributes, which is most evident across different ethnic groups: Black and Hispanic cohorts are less likely to receive ventilation treatments, as well as shorter treatment duration on average. However, there are multiple confounders that may lead to the observed disparity in treatment, which adds the difficulty of identifying intentional discriminations. Hence we call for a close look at causal analysis for a better understanding.
For algorithmic fairness, we evaluate the performance of state-of-the-art machine learning approaches for mortality prediction in terms of AUC-based fairness metrics. Experiment results indicate a strong correlation between mortality rates and fairness: machine learning approaches tend to obtain lower AUC scores on groups with higher mortality rates. Meanwhile, all of the studied mortality predictors are fair in general while IMV-LSTM~\cite{guo2019exploring} performs the best overall across protected groups.
\paragraph{Interactions between interpretability and fairness.} We examine the interaction of interpretability and fairness by drawing connections between feature importance and fairness metrics. Furthermore, we observe substantial disparities in the importance of each demographic feature used for in-mortality prediction across the protected subgroups, which raises a concern whether these demographic features should be used in mortality prediction.

In summary, our main contributions are:
\begin{itemize}
    \item We give quantitative evaluation of various interpretability methods for feature importance estimation on deep learning models in the context of mortality prediction. It is observed that the best performing interpretability method successfully identifies critical features for mortality prediction on various prediction models. Also demographic features are important for prediction.
    \item For dataset bias, we observe treatment disparity among patient groups split by different protected attributes.
    \item For algorithmic fairness, we find that all of the studied mortality predictors are fair in general while the IMV-LSTM model performs the best overall across different protected groups.
    \item We also examine the interaction between interpretability and fairness, and observe disparities of feature importance among demographic subgroups.
\end{itemize}

\section{Related Work}
\subsection{Interpretability Evaluation}
\subsubsection{Aspects of Interpretability of Deep Learning Models}
Due to the complexity of deep learning models, interpretability research has developed diversely, and many methods have been used to interpret how a deep learning model works from various aspects, including: (1) \textit{Feature importance estimation}~\cite{simonyan2013deep,sundararajan2017axiomatic,shrikumar2017learning,ancona2018towards,lundberg2017unified,smilkov2017smoothgrad,castro2009polynomial,strumbelj2010efficient,molnar2020interpretable,suresh2017clinical,zeiler2014visualizing}. For a given data sample, these methods estimate the importance of each individual input feature with respect to a specified output. (2) \textit{Feature interaction attribution}~\cite{tsang2020does,sundararajan2020shapley,janizek2020explaining,sorokina2008detecting,tsang2018detecting,tsang2018neural}. In addition to estimating the importance of individual features, these methods analyze how interactions of feature pairs/groups contribute to predictions. (3) \textit{Neuron/layer attribution}~\cite{dhamdhere2018important,shrikumar2018computationally,leino2018influence,springenberg2015striving,zeiler2014visualizing}. These methods estimate the contribution of specified layers/neurons in the model. (4) \textit{Explanation with high-level concepts}~\cite{kim2018interpretability,ghorbani2019towards,zhou2018interpretable}. These methods interprete deep learning models with human-friendly concepts instead of importance of low-level input features. In this paper, we focus on feature importance estimation due to its importance and the completeness of its evaluation methods.

\subsubsection{Evaluation of Feature Importance Interpretation}
Since feature importance estimation assigns an importance score for each input feature, the evaluation of results is equivalent to the evaluation of binary classification results when the ground truth of feature importance is available, where the label indicates whether the feature is important for the problem. \cite{ismail2020benchmarking} constructs synthetic datasets with feature importance labels for evaluation.~\cite{hardt2020explaining} obtains feature importance labels from both manually constructed tasks and domain experts.~\cite{sanchez2020evaluating} derives importance labels from tasks with graph-valued data with computable ground truths. However, these evaluation methods require the accessibility of ground truth labels, which is hard to fulfill and is usually the problem itself we need to solve in domains such as healthcare.

For evaluation without ground truth, A common strategy to evaluate feature importance estimation is to measure the degradation of model performance with the gradual removal of features estimated to be important.~\cite{samek2016evaluating} pertubates features ranked by importance in test samples and calculate the area over the MoRF curve (AOPC): a higher AOPC means the information disappears faster with feature removal and indicates a better importance estimation.~\cite{hooker2019benchmark} remove features from the entire dataset and retrain the model when obtaining AOPC, which excludes the interference of data distribution shifting.~\cite{ismail2020benchmarking} replace features with known feature distributions for evaluation on synthetic tasks to ensure the consistency of data distribution. In this paper, we utilize the evaluation in~\cite{hooker2019benchmark}.

\subsection{Fairness Evaluation}
\subsubsection{Bias and Fairness in Machine Learning}
With the open access to large-scale datasets and the development of machine learning algorithms, more decisions in the real world are made by machine learning algorithms with or without human's intervention, e.g., job advertisements promoting~\cite{lambrecht2019algorithmic}, facial recognition~\cite{raji2019actionable}, treatment recommendation~\cite{schnabel2016recommendations}, etc. Due to bias in datasets or models, decisions made by machine learning algorithms are prone to be unfair, where an individual or a group is favored compared with the others owing to their inherent traits. One well-known example is the software COMPAS (Correctional Offender Management Profiling for Alternative Sanctions), which was found a bias against African-Americans to assign a higher risk score of recommitting another crime than to Caucasians with the same profile~\cite{dressel2018accuracy}.

Based on the general assumption that the algorithm itself is not coded to be biased, the decision unfairness can be attributed to biases in the data, which is likely to be picked up and amplified by the trained algorithm~\cite{fu2020artificial}. Three major sources of data biases are~\cite{fu2020artificial}: 1) \emph{Biased Labels:} the ground-truth labels for the machine learning algorithms to predict are biased; 2) \emph{Imbalanced representation:} imbalanced representation of different demographic groups occurs when some protected groups are underrepresented with fewer observations in the dataset compared with other groups; 3) \emph{Data Quality Disparity:} data from protected groups might be less complete or accurate during data collecting and processing.
Mostly widely considered traits, such as gender, age, ethnicity, marital status, are considered as protected or sensitive attributes in literature~\cite{mehrabi2019survey}. Fairness has been defined in various ways considering different contexts or applications, two of them are the most widely leveraged for bias detection and correction: \emph{Equal Opportunity}, where the predictions are required to have equal true positive rate across two demographics, and \emph{Equalized Odds}, where an additional constraint is put on the predictor to have equal false positive rate~\cite{hardt2016equality}.
To derive fair decisions with machine learning algorithms, three categories of approaches have been proposed to mitigate biases~\cite{bellamy2018ai,mehrabi2019survey}: 1)\emph{Pre-processing:} the original dataset is transformed so that the underlying discrimination towards some groups is removed~\cite{kamiran2012data}; 2) \emph{In-processing:} either by adding a penalization term in the objective function~\cite{moyer2018invariant} or imposing a fairness-relevant constraint~\cite{singh2019fair}; 3) \emph{Post-processing:} further recompute the results from predictors to improve fairness~\cite{barda2020addressing}.

\subsubsection{Bias and Fairness in MIMIC-III}
With clinical notes~\cite{martinez2020minimax, zhang2020hurtful} or temporal measurements~\cite{chen2019can,cui2020xorder,chen2018my} or both~\cite{chen2020exploring} from MIMIC-III considered, fairness evaluation and bias mitigation have been studied recently for tasks such as mortality prediction~\cite{chen2020exploring,martinez2020minimax, zhang2020hurtful, chen2019can,cui2020xorder, chen2018my}, phenotyping~\cite{chen2020exploring,zhang2020hurtful}, readmission~\cite{chen2019can}, length of stay~\cite{cui2020xorder}, etc. To evaluate data and prediction fairness for the aforementioned healthcare tasks, attributes like ethnicity~\cite{chen2020exploring,martinez2020minimax,zhang2020hurtful,cui2020xorder,chen2018my}, gender~\cite{chen2020exploring,zhang2020hurtful,cui2020xorder}, insurance~\cite{chen2020exploring,zhang2020hurtful}, age~\cite{martinez2020minimax} and language~\cite{zhang2020hurtful}, are considered most often to split patients into different protected groups.

When making medical decisions based on text data like clinical notes, word embeddings, used as machine learning inputs, have been demonstrated to propagate unwanted relationships with regard to different genders, language speakers, ethnicities, and insurance groups~\cite{chen2020exploring,zhang2020hurtful}. With respect to gender and insurance type, differences in accuracy and therefore machine bias has been observed for mortality prediction~\cite{chen2019can}.
To mitigate biases and improve prediction fairness, Chen \emph{et al.} argued that collecting data with adequate sample sizes and predictive variables measures is an effective approach to reduce discrimination without sacrificing accuracy~\cite{chen2018my}. Martinez \emph{et al.} proposed an in-processing approach where the fairness problem is characterized as a multi-objective optimization task, where the risk for each protected group is a separate objective~\cite{martinez2020minimax}. After well-trained machine learning models make predictions, equalized odds post-processing~\cite{chen2020exploring} and updating predictions according to the weighted sum of utility and fairness~\cite{cui2020xorder} were introduced respectively as effective post-processing approaches.

To continue the dataset bias and algorithmic fairness study on MIMIC-IV, we follow previous fairness study work and adopt the following commonly used demographic features as protected attributes: 1) \emph{ethnicity}, 2) \emph{gender}, 3) \emph{marital status}, 4) \emph{Age}, and 5) \emph{insurance type}. For dataset bias, we analyze the average adoption and duration of five types of ventilation treatment on patients from different groups. For algorithmic fairness, we evaluate the performance of state-of-the-art machine learning approaches for mortality prediction in terms of accuracy and fairness.
\subsection{Interactions between Interpretability and Fairness}
Besides accuracy, interpretability and fairness are two important aspects that businesses and researchers should take into consideration when designing, deploying, and maintaining machine learning models~\cite{sharma2019certifai}. It is also well acknowledged that enhancing model interpretability is an important step towards developing fairer ML systems~\cite{chu2020games} since interpretations can help detect and mitigate bias during data collection or labeling~\cite{doshi2017roadmap,lipton2018mythos,du2020fairness}. Given evaluation metrics from the two concepts, demonstrations of performance from different predictive models have been shown in literature to further investigate their interactions~\cite{kleinberg2019simplicity,jabbariempirical,adebayo2016iterative,wadsworth2018achieving,cesaro2019measuring}. When the model's complexity is determined by the number of features and simpler models are more interpretable, curves showing how model fairness is affected by model complexity were studied besides its influence on accuracy~\cite{kleinberg2019simplicity,jabbariempirical}. When the feature importance is leveraged to interpret model predictions, failure of fairness can be identified by detecting whether the feature has a larger effect than it should have~\cite{adebayo2016iterative,wadsworth2018achieving}. For instance, Adebayo \emph{et al.} showed that gender is of low importance among all studied demographic features in a bank's credit limit model, which indicates that the bank's algorithm is not overly dependent on gender in making credit limit determinations~\cite{adebayo2016iterative}. Recently, connections between interpretability and fairness were quantitatively studied by comparing fairness measures and feature importance measure: there is a direct relation between SHAP value difference and equality of opportunity after removing bias with reweighing techniques and measuring feature importance with SHAP on Adult, German, Default and COMPAS datasets~\cite{cesaro2019measuring}. Given mortality predictions made by state-of-the-art models on MIMIC-IV, we study the connections between feature importance induced by different interpretation approaches and the fairness measures in this paper.

\section{MIMIC-IV Dataset}

In this section, we describe the following preprocessing steps of the MIMIC-IV dataset: cohort selection, feature selection, and data cleaning. We also report the distributions of demographic, admission and comorbidity variables of the preprocessed dataset.

\subsection{Dataset Description}
MIMIC-IV~\cite{johnson2020mimiciv,goldberger2000physionet} is a publicly available database of patients admitted to the Beth Israel Deaconess Medical Center (BIDMC) in Boston, MA, USA. It contains de-identified data of 383,220 patients admitted to an intensive care unit (ICU) or the emergency department (ED) between 2008 - 2019. Till the day when we finished all experiments, the latest version of MIMIC-IV is v0.4 and only provides public access to the electronic health record data of 50,048 patients admitted to the ICU, which is sourced from the clinical information system MetaVision at the BIDMC. Therefore, we design the following data preprocessing procedures for the ICU data part of MIMIC-IV.

\subsection{Preprocessing}

\subsubsection{Cohort Selection}
Following the common practice in~\cite{PURUSHOTHAM2018112,wang2020mimic}, we select ICU stays satisfying the following criteria as the cohort: (1) the patient is at least 15 years old at the time of ICU admission; (2) the ICU stay is the first known ICU stay of the patient; (3) the total duration of ICU stay is between 12 hours and 10 days. After the cohort selection, we collect 45,768 ICU stays as the cohort. According to the cohort selection criterion (2), each ICU stay corresponds to one unique patient and one unique hospital admission.

\subsubsection{Data Cleaning}
We follow the same data cleaning procedure in~\cite{PURUSHOTHAM2018112} to handle: (1) Inconsistent units. We convert features with multiple units to their major unit. (2) Multiple recordings at the same time. We use the average value for numerical features and the first appearing value for categorical features. (3) Range of feature values. We use the median of the range as the value of the feature.

\subsubsection{Feature Selection}
We select 164 features from the following groups:
\begin{itemize}
    \item Electronic healthcare records (EHR). We modify the feature list used in~\cite{PURUSHOTHAM2018112} and extract 122 features after removing features that are no longer available in MIMIC-IV.
    \item Demographic features. We extract 5 from patients' demographic information.
    \item Admission features. We extract 4 from admission records.
    \item Comorbidity features. We extract binary flags of 33 types of comorbidity using patients' ICD codes as comorbidity features.
\end{itemize}

We provide a detailed list of all selected features in \tref{tab:full-list} in Appendix.

\subsubsection{Data Filtering, Truncation, Aggregation and Imputation}
\textbf{Data Filtering} After specifying the list of features, we further filter ICU stays from the cohort and only keep those that have records of selected EHR features for at least 24 hours and at most 10 days, starting from the first record within 6 hours prior to ICU admission time. We have 43005 ICU stays after the filtering.

Other works such as~\cite{wang2020mimic} extract the first 30-hour data and drop the data from the last 6 hours to avoid information leakage of positive mortality labels to features measured within 6 hours prior to deathtime. We find that most (96.02\%) of the patients with positive in-hospital mortality labels have measurement for over 30 hours prior to their deathtime, thus we omit this processing step.

\textbf{Truncation} For each ICU stay, we only keep the data of the first 24 hours, starting from the first record within 6 hours prior to its ICU admission time.

\textbf{Aggregation} For each ICU stay, we aggregate its records hourly by taking the average of multiple records within the same hourly time window.

\textbf{Imputation} We perform forward and backward imputation to fill missing values. For cases where certain features of some patients are completely missing, we fill with mean values of corresponding features in the training set.

\subsection{Dataset Summary}
\begin{wraptable}{r}{0.58\textwidth}
    \centering
    \caption{Differences between preprocessed MIMIC-III in~\cite{PURUSHOTHAM2018112} and preprocessed MIMIC-IV.}
    \label{tab:one}
    \resizebox{0.58\textwidth}{!}{
        \begin{tabular}{@{}ccc@{}}
            \toprule
                                    & MIMIC-III & MIMIC-IV (this work) \\ \midrule
            \# Samples              & 35627          & 43005                     \\
            \# Temporal Features    & 135          & 122                     \\
            \# Demographic Features & 1          & 5                     \\
            \# Admission Features   & 1          & 4                     \\
            \# Comorbidity Features & 3          & 33                     \\\bottomrule
        \end{tabular}
    }
\end{wraptable}
After all preprocessing steps, we obtain features of the shape $(N, T, F)$, where $N=43005$ is the number of ICU stays (data samples), $T=24$ is the number of time steps with 1-hour step size, and $F=164$ is the total number of features. We also process the data into the tabular form $(N, F^\prime)$ by replacing sequential EHR features with the summary over time steps including minimum, maximum, and mean values (for the urinary\_output\_sum feature we have summation in addition), where $F^\prime = 409$.

We show the distribution of demographic features, admission features and comorbidity features grouped by patients' in-hospital mortality status in \tref{tab:dist-mor} in Appendix. We also demonstrate differences between the preprocessed MIMIC-IV data in this work and the preprocessed MIMIC-III data from ~\cite{PURUSHOTHAM2018112} in Table \ref{tab:one}.

\section{Interpretability Evaluation}
In this section, we evaluate the performance of various feature importance interpretability methods on multiple models for the in-hospital mortality prediction task. We describe the task, models, interpretability methods, and the evaluation method in detail and report the evaluation results.

\subsection{Task Description}
Mortality prediction is one primary outcome of high interest of hospital admissions, and is widely considered in other benchmark works~\cite{PURUSHOTHAM2018112, harutyunyan2019multitask, sjoding2019democratizing, wang2020mimic}. We use the in-hospital mortality prediction task to train different models and evaluate the performance of various interpretability methods.

We formulate the in-hospital mortality prediction task as a binary classification task. Given the observed sequence of features $\mX\in \mathbb{R}^{T\times F}$ of one patient (or its summary $\vx\in \mathbb{R}^{F}$, depending on the model), the model gives the probability that the patient dies during his/her hospital admission after being admitted to ICU. In MIMIC-IV, a patient has in-hospital mortality if and only if his/her deathtime exists in the \textbf{mimic\_core.admissions} table. We randomly divide 60\% data for training, 20\% for validation and 20\% for test.

\subsection{Models}\label{subsec:models}
We consider following models: (1) \textbf{AutoInt}~\cite{song2019autoint}. A model that learns feature interaction automatically via self-attentive neural networks. (2) \textbf{LSTM}~\cite{hochreiter1997long}. Long short-term memory recurrent neural network, which is a common baseline for sequence learning tasks. (3) \textbf{TCN}~\cite{bai2018empirical}. Temporal convolutional networks, which outperform canonical recurrent networks across various tasks and datasets. (4) \textbf{Transformer}~\cite{vaswani2017attention}. A network architecture based solely on attention mechanisms. Here we only adopt its encoder part for the classification task. (5) \textbf{IMVLSTM}~\cite{guo2019exploring}. An interpretable model that jointly learns network parameters, variable and temporal importance, and gives inherent feature importance interpretation. We use sequence data as input for (2)-(5), and the summary of sequence data as input for (1) since AutoInt only processes tabular data in its original implementation.

We use the area under the precision-recall curve (AUPRC) and the area under the receiver operating characteristic curve (AUROC) as metrics for binary classification. The performance of all models considered in this work is shown in \tref{tab:interpretability-perf}.

\begin{table}[htbp]
    \centering
    \caption{Classification performance of all considered deep models.}
    \label{tab:interpretability-perf}
    \resizebox{0.87\textwidth}{!}{
    \begin{tabular}{cccccccccc}
        \toprule
        \multicolumn{2}{c}{AutoInt} & \multicolumn{2}{c}{LSTM} & \multicolumn{2}{c}{TCN} & \multicolumn{2}{c}{Transformer} & \multicolumn{2}{c}{IMVLSTM}                                         \\ \cmidrule(l){1-10}
        AUPRC                       & AUROC                    & AUPRC                   & AUROC                           & AUPRC                       & AUROC & AUPRC & AUROC & AUPRC & AUROC \\
        \midrule
        0.508                       & 0.901                    & 0.660                   & 0.938                           & 0.666                       & 0.928 & 0.686 & 0.939 & 0.769 & 0.955 \\
        \bottomrule
    \end{tabular}}
\end{table}

\subsection{Interpretability Methods}
Interpretation of deep learning models is still a rapidly developing area and contains various aspects. In this work, we focus on the interpretation of feature importance, which estimates the importance of single features for a given model on a specific task. Estimation of feature importance helps improve the model, builds trust in prediction and isolates undesirable behavior~\cite{hooker2019benchmark}. In addition, recent works~\cite{samek2016evaluating,hooker2019benchmark,ismail2020benchmarking} have developed methods for evaluating feature performance estimation without access to the ground truth of feature importance, which fits scenarios in healthcare domains well: ground-truth feature importance for healthcare applications is either the problem we need to solve itself or requires extraction from a huge amount of domain knowledge. Therefore, we choose the interpretation of feature importance as the target aspect for evaluating interpretability methods.

Formally, given a function $M: \mathbb{R}^{d_{in}}\rightarrow \mathbb{R}^{d_{out}}$ and the input (flattened) feature vector $\vx\in\mathbb{R}^{d_{in}}$, the interpretation of feature importance gives a non-negative score $\vs(\vx)\in\mathbb{R}^{d_{in}}$, where $s(\vx)_i$ is the importance of $x_i$ to $M(\vx)$.

We select the following interpretability methods to compare their feature importance estimation results. Notice that some interpretability methods give signed scores (or "attributions"), where signs reflect positive/negative contributions of features to the output, and we use the absolute values of signed scores as importance scores. For methods requiring a baseline input vector, unless otherwise specified, we follow the method in\cite{ismail2020benchmarking} and randomly sample $\vx^\prime\in\mathbb{R}^{d_{in}}$, where $x^\prime_i \sim \mathcal{U}[0, 1]$.

\paragraph{(1) Graidient based methods.}
\begin{itemize}
    \item \textbf{Saliency}~\cite{simonyan2013deep}. Saliency returns the gradients with respect to inputs as feature importance: $\vs(\vx) = \frac{\partial M(\vx)}{\partial \vx}$. By taking the first-order Taylor expansion of the neural network at the input, $M(\vx)\approx (\frac{\partial M(\vx)}{\partial \vx})^{\intercal}\vx + b$, which is a linear approximation of the network, the gradient $\frac{\partial M(\vx)}{\partial x_i}=s(\vx)_i$ is the coefficient of the $i$-th feature.
    \item \textbf{IntegratedGradients}~\cite{sundararajan2017axiomatic}. IntegratedGradients assigns an importance score to each input feature by approximating the integral of gradients of the model’s output with respect to the inputs along the path (straight line) from given baselines to inputs, i.e.
          \begin{equation}
              \mathrm{IntegratedGradients}(\vx)_i = (x_i - x^\prime_i)\times \int_{\alpha=0}^1 \frac{\partial M(\vx^\prime + \alpha(\vx - \vx^\prime))}{\partial x_i} d\alpha,
          \end{equation}
          where $\vx^\prime$ is the baseline.
    \item \textbf{DeepLift}~\cite{shrikumar2017learning,ancona2018towards}. DeepLift decomposes the output prediction of a neural network on a specific input by backpropagating the contributions of all neurons in the network to every feature of the input.
          \begin{equation}
              \mathrm{DeepLift}(\vx)_i = (x_i - x^\prime_i) \times \frac{\partial^g M(\vx)}{\partial x_i}, g(z_t) = \frac{f_t(z_t) - f_t(z_t^\prime)}{z_t - z_t^\prime},
          \end{equation}
          where $\frac{\partial^g M(\vx)}{\partial x_i} = \sum_{p\in P_{io}}(\prod_{(s,t)\in p} w_{ts} \prod_{(s,t)\in p} g(z_t))$. $P_{io}$ is the set of all paths from the $i$-th input feature to the output neuron in the network. $(s,t)$ is a pair of connected neurons in path $p$. Each neuron $t$ contains a linear transformation $z_t = \sum_{q\in Pa(t)}w_{tq}o_q + b_t$ followed by a nonlinear mapping $o_t = f(z_t)$.
    \item \textbf{GradientShap}~\cite{lundberg2017unified}. GradientShap approximates SHAP (SHapley Additive exPlanations) values by computing the expectations of gradients by randomly sampling from the distribution of baselines. It first adds white noise to each input sample and selects a random baseline from a given distribution, then selects a random point along the path between the baseline and the input with noise, and computes the gradient of outputs with respect to the random point. The procedure is repeated for multiple times to approximate the expected values of gradients $E(\frac{\partial M(\vx)}{\partial \vx})$. The final SHAP value for the $i$-th feature is $E(\frac{\partial M(\vx)}{\partial x_i}) \times (x_i - x^\prime_i)$.
    \item \textbf{DeepLiftShap}~\cite{lundberg2017unified}. It extends DeepLift algorithm and approximates SHAP values using DeepLift. For each input, it samples baselines from a given distribution and computes the DeepLift score for each input-baseline pair and averages the resulting scores per input example as the output.
    \item \textbf{SaliencyNoiseTunnel}~\cite{smilkov2017smoothgrad}. SaliencyNoiseTunnel adds Gaussian noise to the input sample and averages the calculated attributions using Saliency method as the output.
\end{itemize}

\paragraph{(2) Perturbation based methods.}
\begin{itemize}
    \item \textbf{ShapleySampling}~\cite{castro2009polynomial,strumbelj2010efficient}. Shapley value gives attribution scores by taking each permutation of the input features and adding them one-by-one to a given value. Since the computation complexity is extremely high for large numbers of features, ShapleySampling takes some random permutations of the input features and averages the marginal contribution of features.
    \item \textbf{FeaturePermutation}~\cite{molnar2020interpretable}. FeaturePermutation permutes the input feature values randomly within a batch and computes the difference between original and shuffled outputs as the result.
    \item \textbf{FeatureAblation}~\cite{suresh2017clinical}. FeatureAblation replaces each input feature with a given baseline value and computes the difference in output as the result.
    \item \textbf{Occlusion}~\cite{zeiler2014visualizing}. Occlusion replaces each contiguous rectangular region with a given baseline and computing the difference in output as the result.
    \item \textbf{ArchDetect}~\cite{tsang2020does}. It utilizes the discrete interpretation of partial derivatives. While the original paper considers both single features and feature pairs, we here only apply it to single features, since the evaluation method in this work is designed for single feature importance only. In the single feature case, the importance score of the $i$-th feature is
          \begin{equation}
              \mathrm{ArchDetect}(\vx)_i = \left(\frac{M(\vx_{\{i\}} + \vx^\prime_{\backslash\{i\}}) - M(\vx^\prime_{\{i\}})}{x_i - x^\prime_i}\right)^2,
          \end{equation}
          where \begin{equation}
              (\vx_{\mathcal{I}})_i = \left\{
              \begin{array}{cc}
                  x_i , & ~\mathrm{if}~i\in\mathcal{I};  \\
                  0   , & ~\mathrm{otherwise}          . \\
              \end{array}
              \right.
          \end{equation} Here we select $\vx^\prime = \vzero\in\mathbb{R}^{d_{in}}$.
\end{itemize}

\paragraph{(3) Glassbox interpretation.}
If the model's architecture provides feature importance scores directly as a part of the output of the model, such as the attention score of each feature, we call this interpretation as "Glassbox" and regard it as an extra baseline.

\paragraph{(4) Random baseline.}
As a baseline, we randomly shuffle all features as the feature importance ranking.

For models in Section \ref{subsec:models}, AutoInt maps categorical features to embeddings using learnable dictionaries and has no gradient on categorical features, thus gradient based methods are not applicable. Only IMVLSTM model has Glassbox interpretation.

\subsection{Evaluation Method}
Since acquiring the ground-truth feature importance is challenging for mortality prediction tasks, we evaluate one feature importance estimation by gradually dropping most important features it gives at certain ratios from the dataset and observe the degradation of the model's performance. The larger the degradation is, the better the estimation is, since it identifies the features most helpful for the model on the task.

More specifically, we use \textbf{ROAR} (remove and retrain) proposed in~\cite{hooker2019benchmark} for evaluation. For each interpretability method, we replace the most important features of certain fractions of each data sample with a fixed uninformative value. We conduct this in both training and test sets. Then we retrain the model with the modified training set and evaluate its classification performance on the modified test set. By retraining the model on datasets with features removed, ROAR ensures that train and test data comes from a similar distribution and reduces the impact on the model's performance of data distribution discrepancy, so that the degradation of performance is caused by the removal of information instead of the shift of data distribution.

For sequence input $\mX\in\mathbb{R}^{T\times F}$, we flatten it and give feature importance scores for all $T\times F$ features. For the $i$-th feature, we use its mean value in the training set as its uninformative value. We evaluate each interpretability method with feature drop ratios $10\%,20\%,\dots,100\%$ and plot the curve of model performance with respect to feature drop ratio for each model.

\subsection{Results}

\subsubsection{Evaluation of Interpretability Methods}
\label{subsubsec:eval-of-interp}
\begin{figure}
    \centering
    \begin{subfigure}[b]{0.45\textwidth}
        \centering
        \includegraphics[width=\textwidth]{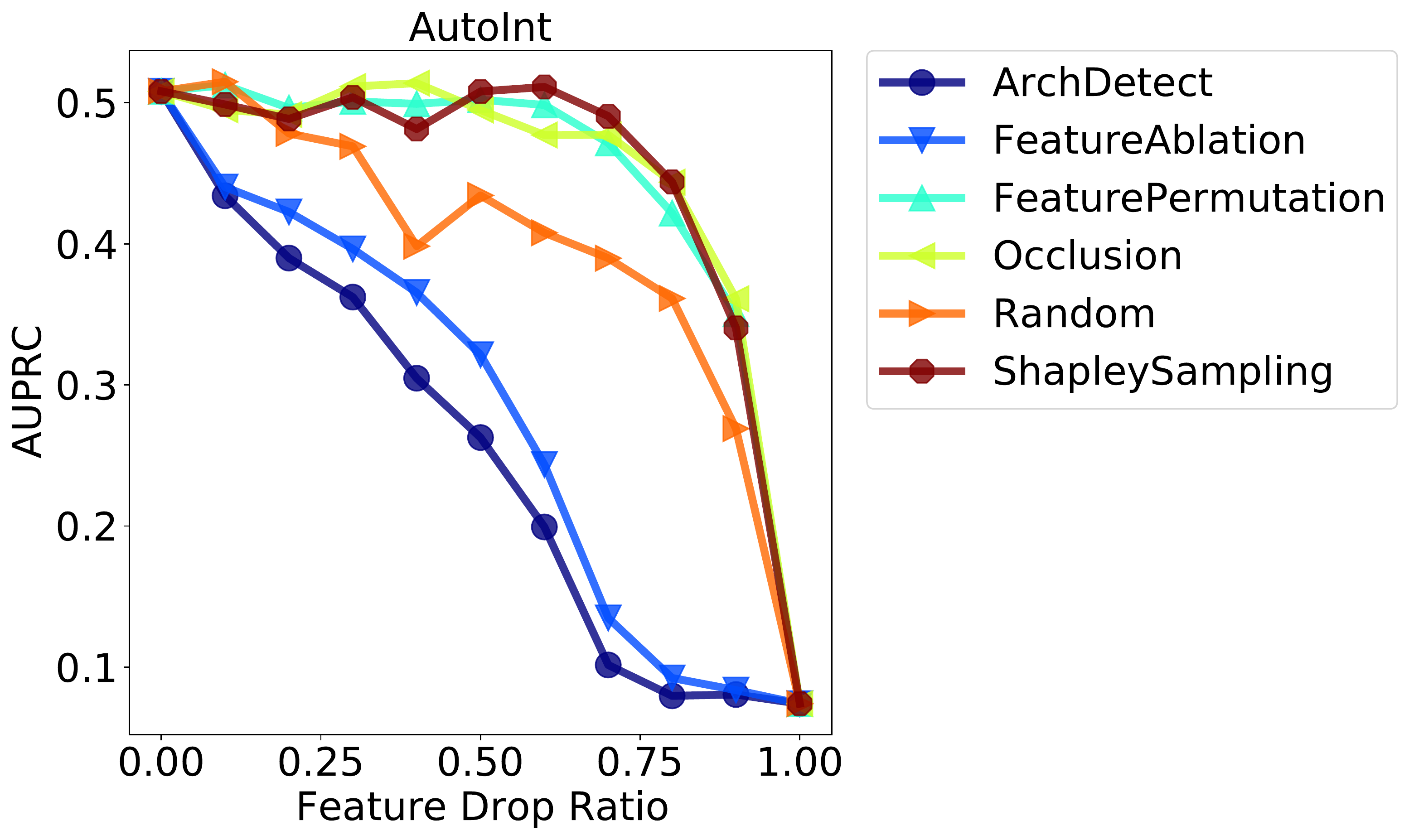}
    \end{subfigure}
    \hfill
    \begin{subfigure}[b]{0.45\textwidth}
        \centering
        \includegraphics[width=\textwidth]{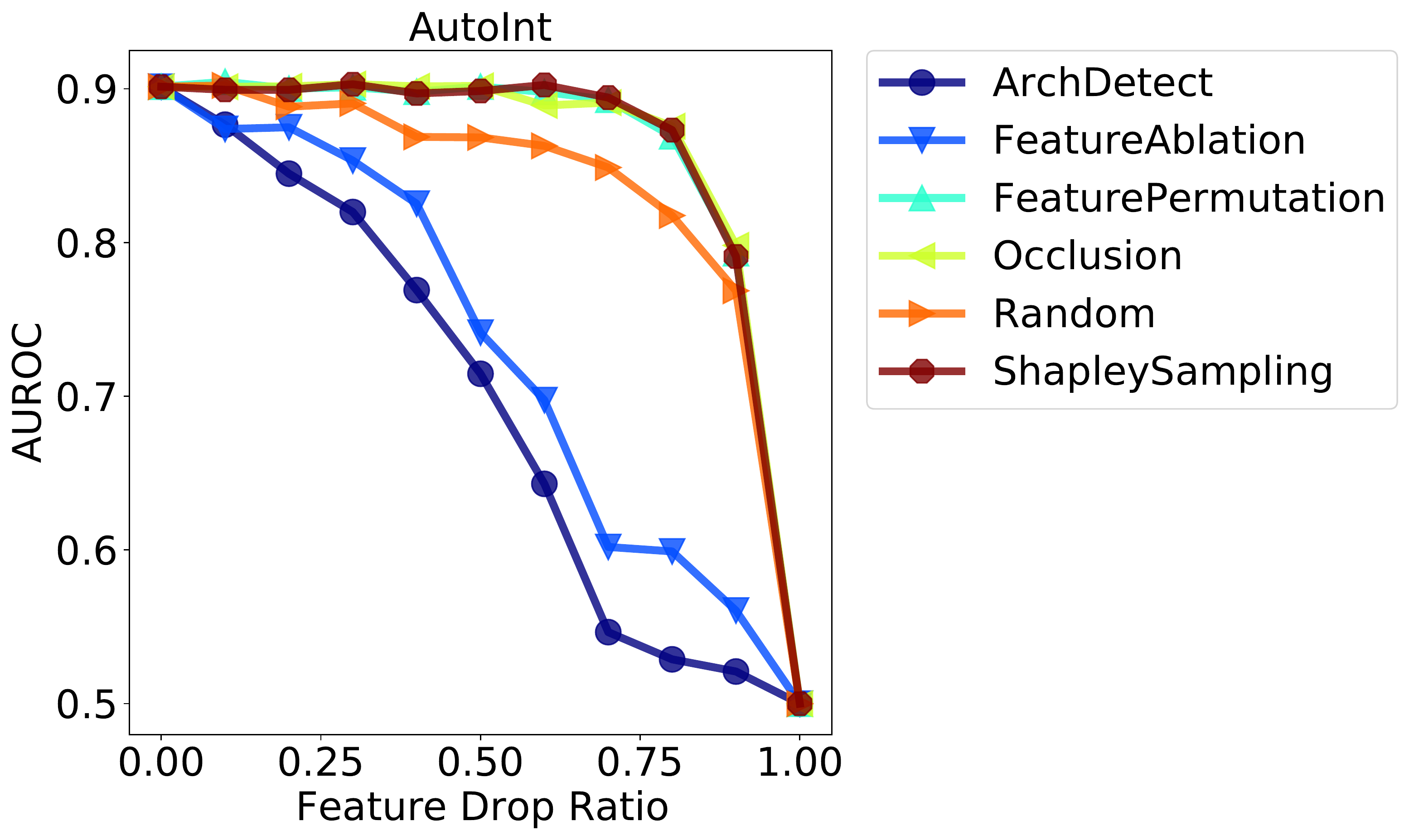}
    \end{subfigure}
    \hfill
    \begin{subfigure}[b]{0.45\textwidth}
        \centering
        \includegraphics[width=\textwidth]{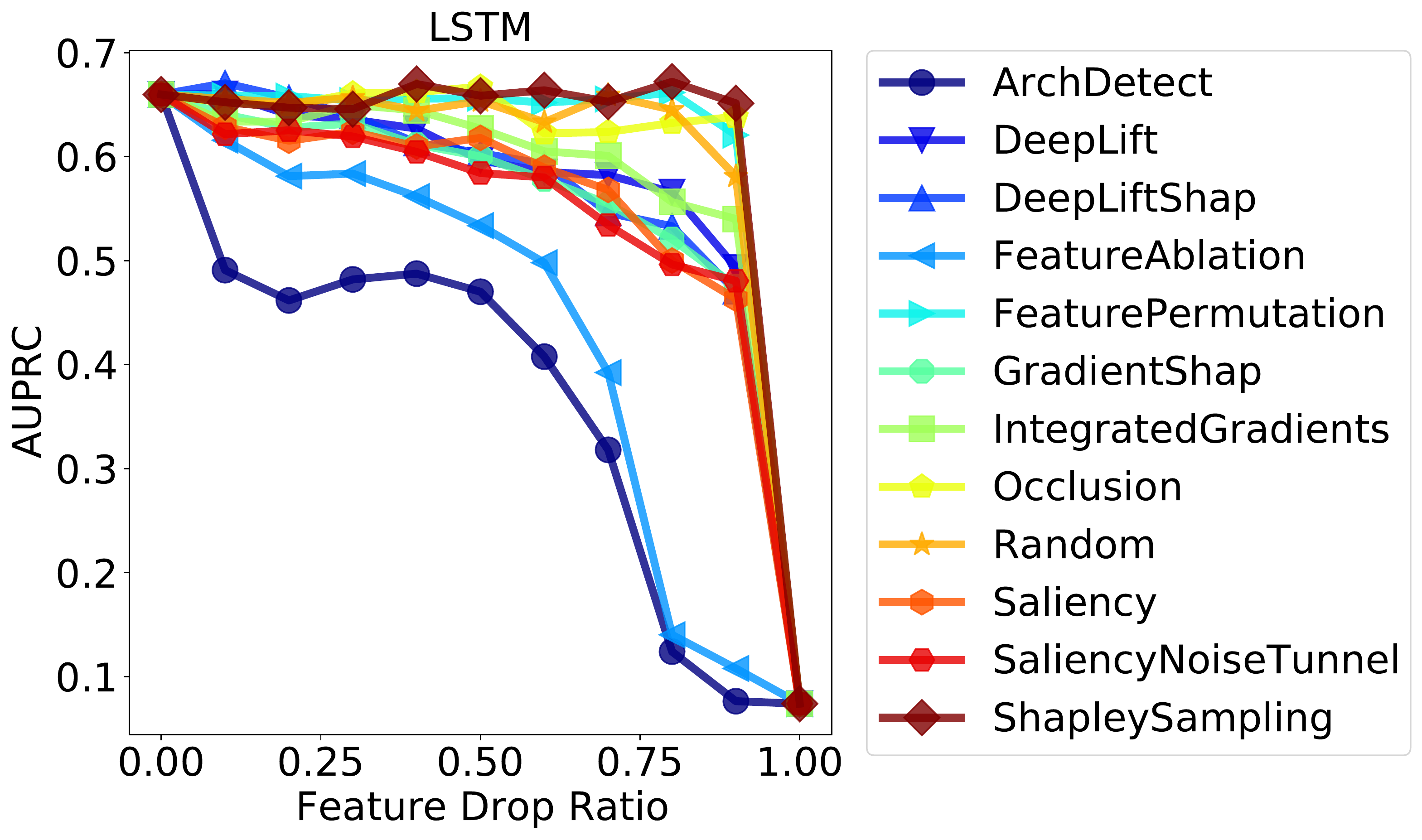}
    \end{subfigure}
    \hfill
    \begin{subfigure}[b]{0.45\textwidth}
        \centering
        \includegraphics[width=\textwidth]{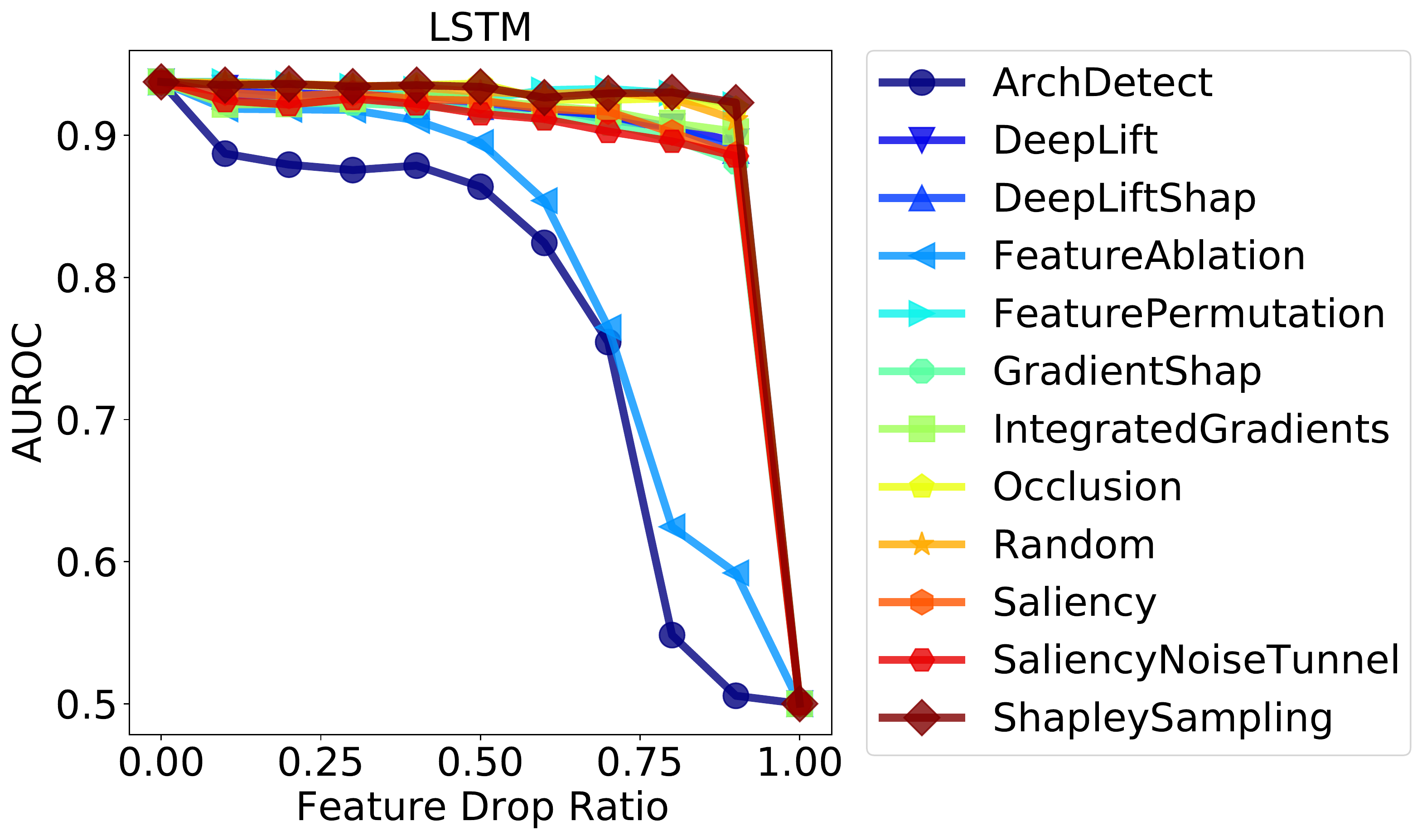}
    \end{subfigure}
    \hfill
    \begin{subfigure}[b]{0.45\textwidth}
        \centering
        \includegraphics[width=\textwidth]{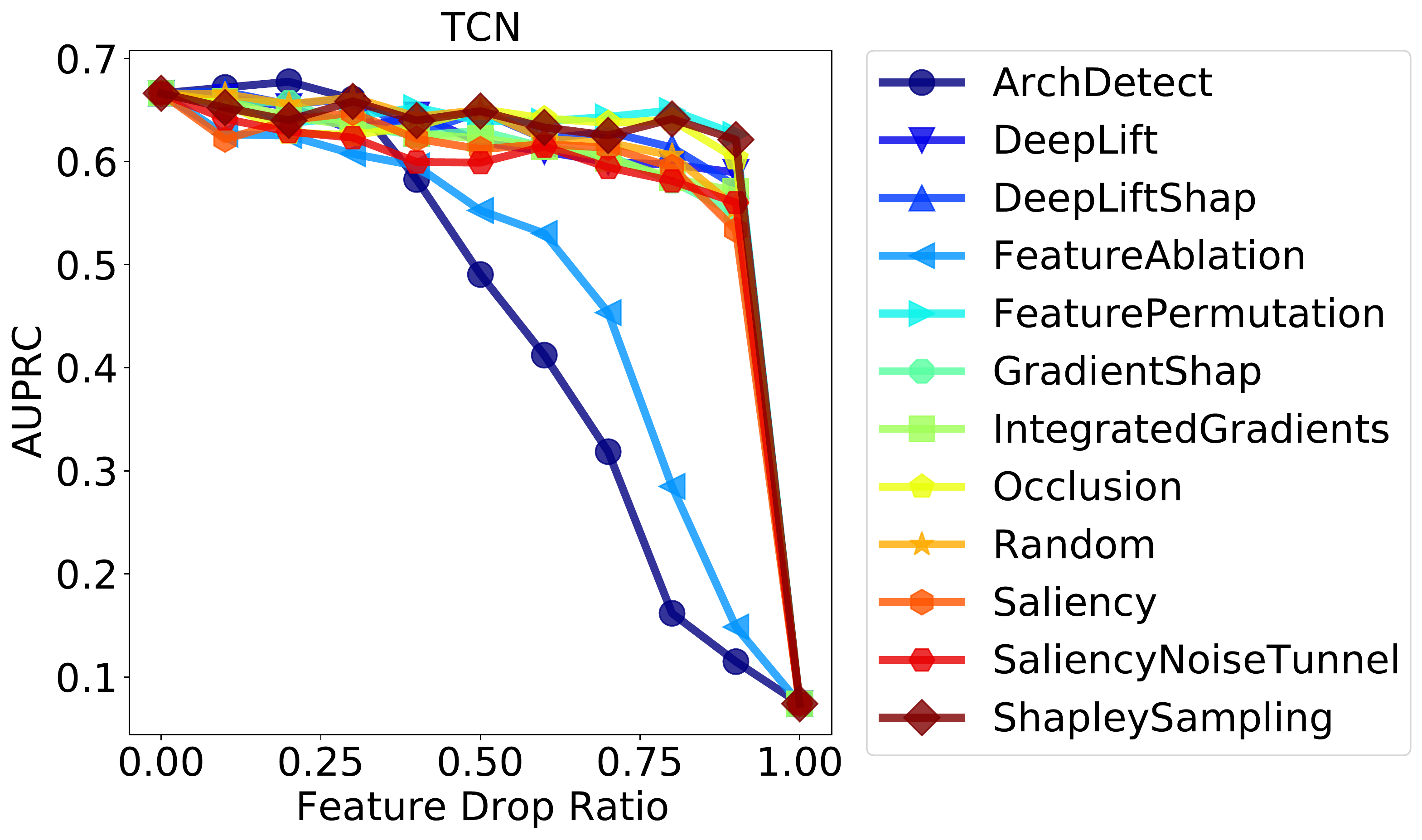}
    \end{subfigure}
    \hfill
    \begin{subfigure}[b]{0.45\textwidth}
        \centering
        \includegraphics[width=\textwidth]{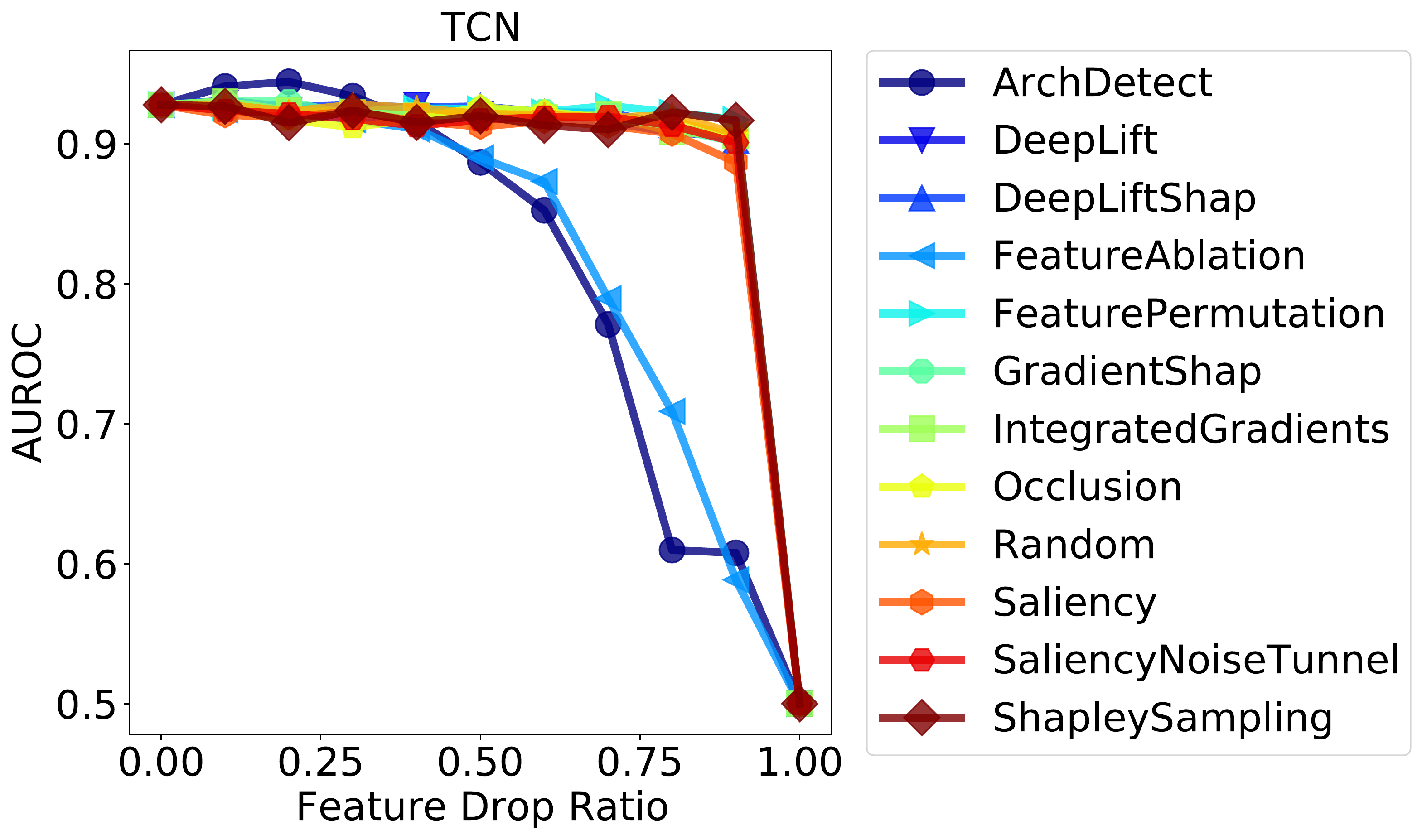}
    \end{subfigure}
    \hfill
    \begin{subfigure}[b]{0.45\textwidth}
        \centering
        \includegraphics[width=\textwidth]{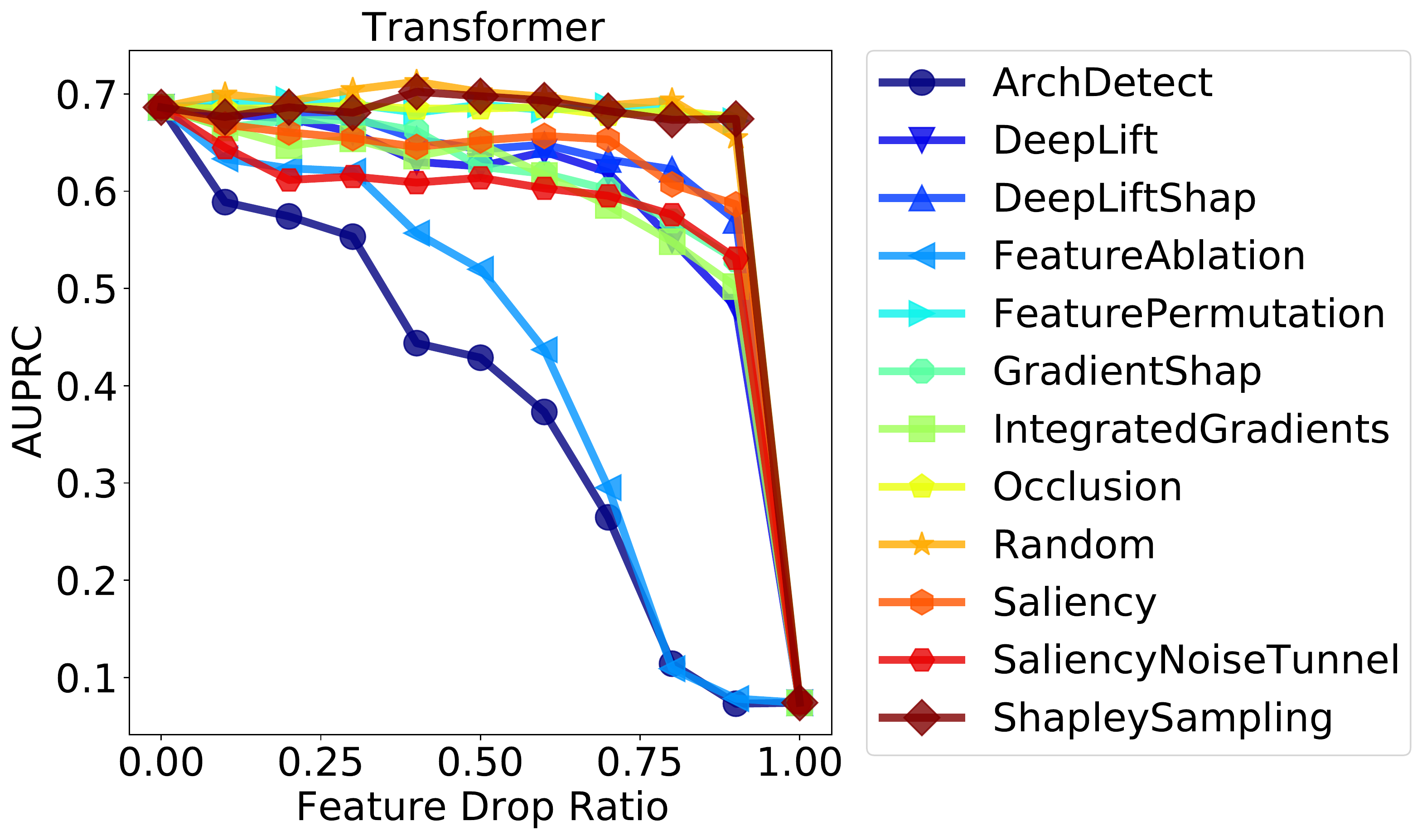}
    \end{subfigure}
    \hfill
    \begin{subfigure}[b]{0.45\textwidth}
        \centering
        \includegraphics[width=\textwidth]{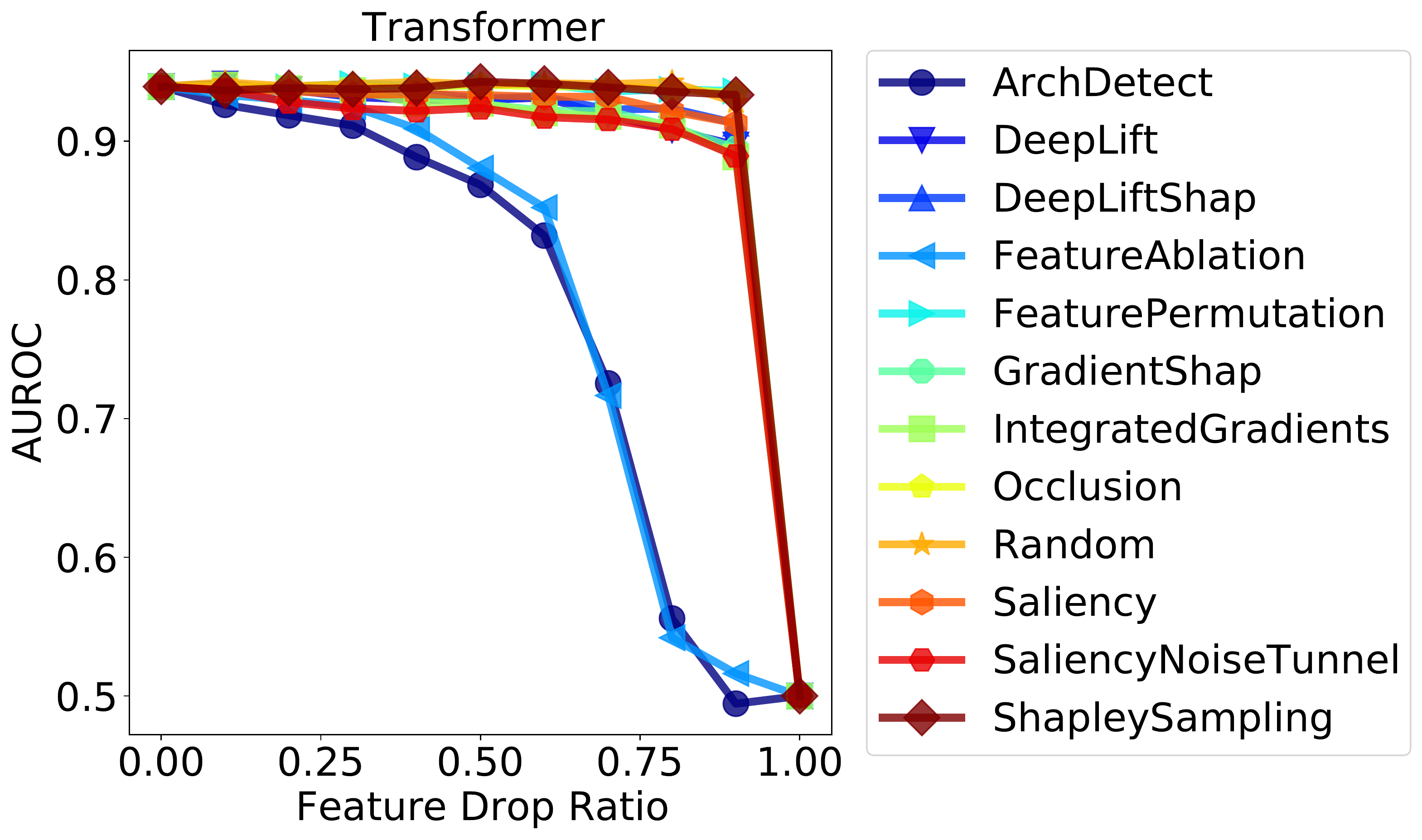}
    \end{subfigure}
    \hfill
    \begin{subfigure}[b]{0.45\textwidth}
        \centering
        \includegraphics[width=\textwidth]{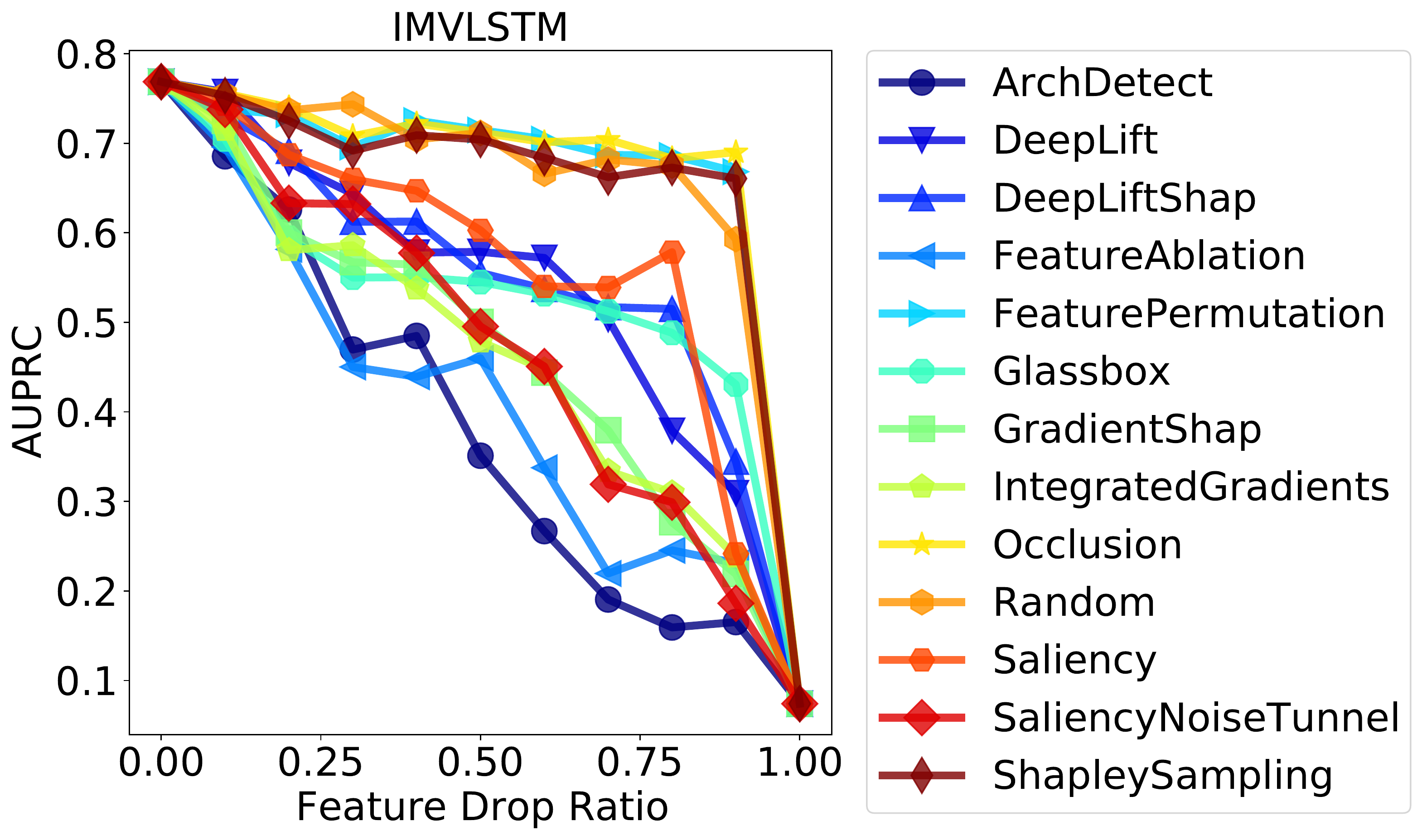}
    \end{subfigure}
    \hfill
    \begin{subfigure}[b]{0.45\textwidth}
        \centering
        \includegraphics[width=\textwidth]{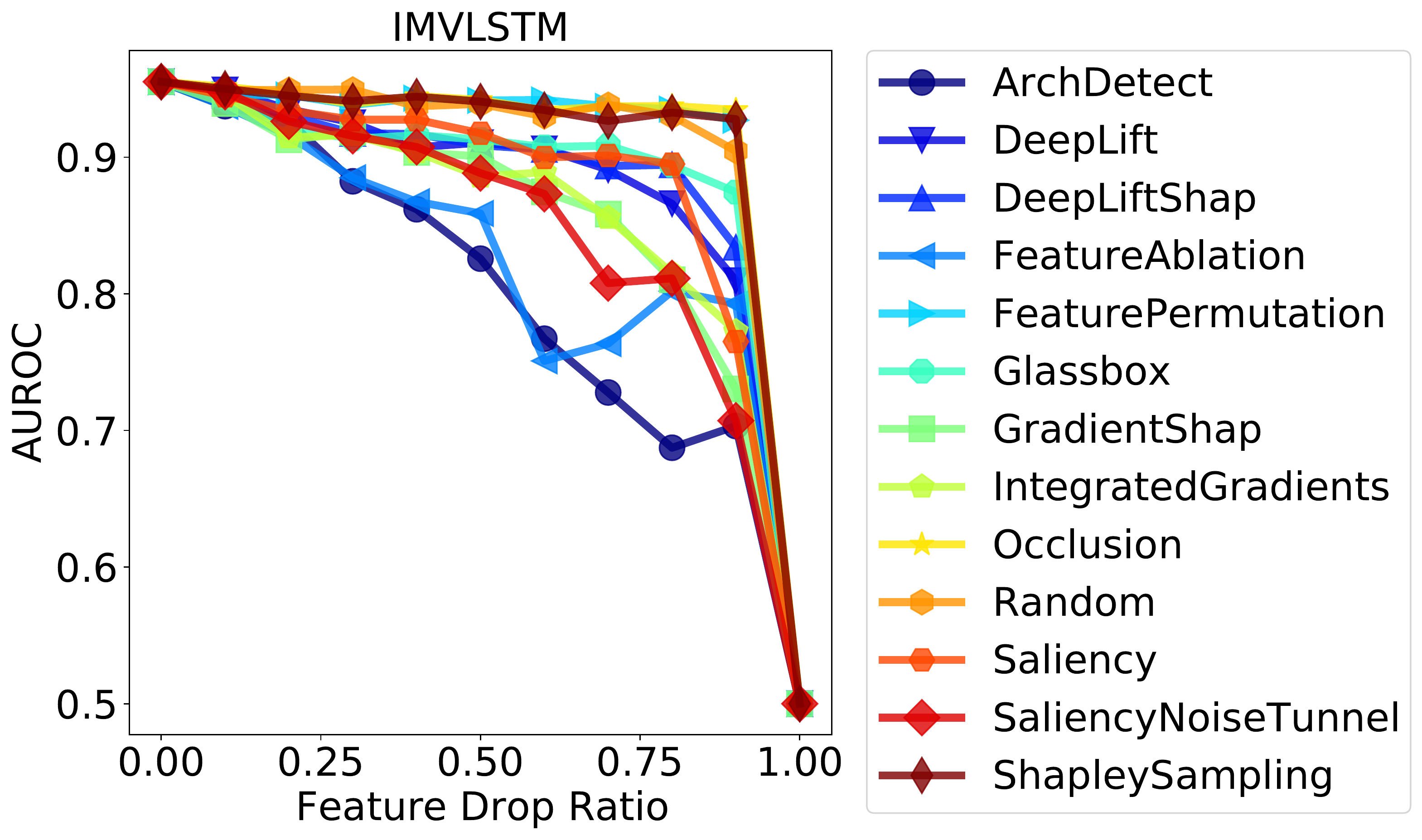}
    \end{subfigure}
    \caption{Curves of performance metric w.r.t feature drop ratio.}
    \label{fig:interpretability-curve}
\end{figure}

\fref{fig:interpretability-curve} shows the curves of model performance (measured with AUPRC and AUROC respectively) with respect to the feature drop ratio of different interpretability methods for each model. \tref{tab:interpretability-auc} gives the quantitative results of area under the curve (AUC). A lower value of AUC means that the performance curve drops faster with the increase of feature drop ratio, thus indicates that the interpretability method gives a better ranking of feature importance.

\begin{table}[htbp]
    \centering
    \caption{Area under the curve (AUC) of interpretability methods for each model and each classification performance metric evaluated using ROAR. AUC is measured for two prediction metrics (AURPC and AUROC) respectively. Lower AUC indicates more rapid prediction performance drop and better feature importance interpretation.}
    \label{tab:interpretability-auc}
    \resizebox{\textwidth}{!}{
        \begin{tabular}{ccccccccccc}
            \toprule
            \multirow{2}{*}{Interpreters} & \multicolumn{2}{c}{AutoInt} & \multicolumn{2}{c}{LSTM} & \multicolumn{2}{c}{TCN} & \multicolumn{2}{c}{Transformer} & \multicolumn{2}{c}{IMVLSTM}                                                                                      \\ \cmidrule(l){2-11}
                                          & AUPRC                       & AUROC                    & AUPRC                   & AUROC                           & AUPRC                       & AUROC          & AUPRC          & AUROC          & AUPRC          & AUROC          \\
            \midrule
            Random                        & 0.401                       & 0.842                    & 0.615                   & 0.909                           & 0.605                       & 0.901          & 0.662          & 0.918          & 0.669          & 0.915          \\
            Glassbox                      & \xmark                      & \xmark                   & \xmark                  & \xmark                          & \xmark                      & \xmark         & \xmark         & \xmark         & 0.533          & 0.892          \\
            Saliency                      & \xmark                      & \xmark                   & 0.558                   & 0.898                           & 0.587                       & 0.893          & 0.616          & 0.909          & 0.566          & 0.884          \\
            IntegratedGradients           & \xmark                      & \xmark                   & 0.586                   & 0.899                           & 0.593                       & 0.899          & 0.588          & 0.903          & 0.465          & 0.863          \\
            DeepLift                      & \xmark                      & \xmark                   & 0.575                   & 0.900                           & 0.598                       & 0.898          & 0.594          & 0.905          & 0.542          & 0.883          \\
            GradientShap                  & \xmark                      & \xmark                   & 0.561                   & 0.893                           & 0.592                       & 0.899          & 0.600          & 0.904          & 0.470          & 0.858          \\
            DeepLiftShap                  & \xmark                      & \xmark                   & 0.569                   & 0.897                           & 0.607                       & 0.901          & 0.619          & 0.909          & 0.554          & 0.887          \\
            SaliencyNoiseTunnel           & \xmark                      & \xmark                   & 0.551                   & 0.892                           & 0.581                       & 0.896          & 0.578          & 0.899          & 0.475          & 0.851          \\
            ShapleySampling               & 0.456                       & 0.866                    & 0.628                   & 0.910                           & 0.613                       & 0.898          & 0.655          & 0.916          & 0.668          & 0.917          \\
            FeaturePermutation            & 0.454                       & 0.866                    & 0.624                   & 0.910                           & 0.616                       & 0.903          & 0.655          & 0.917          & 0.677          & 0.918          \\
            FeatureAblation               & 0.279                       & 0.733                    & 0.438                   & 0.811                           & 0.479                       & 0.824          & 0.425          & 0.792          & 0.408          & 0.830          \\
            Occlusion                     & 0.456                       & 0.866                    & 0.617                   & 0.909                           & 0.609                       & 0.898          & 0.653          & 0.917          & 0.684          & 0.920          \\
            ArchDetect                    & \textbf{0.251}              & \textbf{0.696}           & \textbf{0.369}          & \textbf{0.774}                  & \textbf{0.446}              & \textbf{0.818} & \textbf{0.379} & \textbf{0.784} & \textbf{0.382} & \textbf{0.805} \\
            \bottomrule
        \end{tabular}}
\end{table}


We have the following observations: (1) \textbf{ArchDetect gives the best performing feature importance estimation overall.} From \fref{fig:interpretability-curve}, we observe that the curve of ArchDetect drops the fastest for all models on both metrics. Quantitative results in \tref{tab:interpretability-auc} also show that ArchDetect has the lowest AUC. Therefore, for the in-hospital mortality task, the feature importance ranking given by ArchDetect is the most reasonable one among results of all interpretability methods considered in this work. (2) \textbf{Gradient based methods perform well on LSTM, Transformer and IMVLSTM models, but are no better than a random guess on TCN.} AUC of both metrics of gradient based methods is significantly lower than that of random guessing for LSTM, Transformer and IMVLSTM. But for TCN, even the best performing gradient based method SaliencyNoiseTunnel has AUC close to random guessing (0.581 vs 0.605 for AUPRC and 0.896 vs 0.901 for AUROC). (3) \textbf{Attention scores are not necessarily the best estimation of feature importance.} In IMVLSTM, the Glassbox baseline utilizes attention scores the model gives as an estimation of feature importance. Although it outperforms the random guessing baseline, it is not among the best interpretation methods and is inferior to methods such as ArchDetect, IntegratedGradients and GradientShap. Similar observations also exist in the natural language processing domain~\cite{jain2019attention,grimsley-etal-2020-attention}, where attention weights largely do not correlate with feature importance.

\subsubsection{Identified Important Features}
\begin{wrapfigure}{r}{0.3\textwidth}
    \centering
    \includegraphics[width=\linewidth]{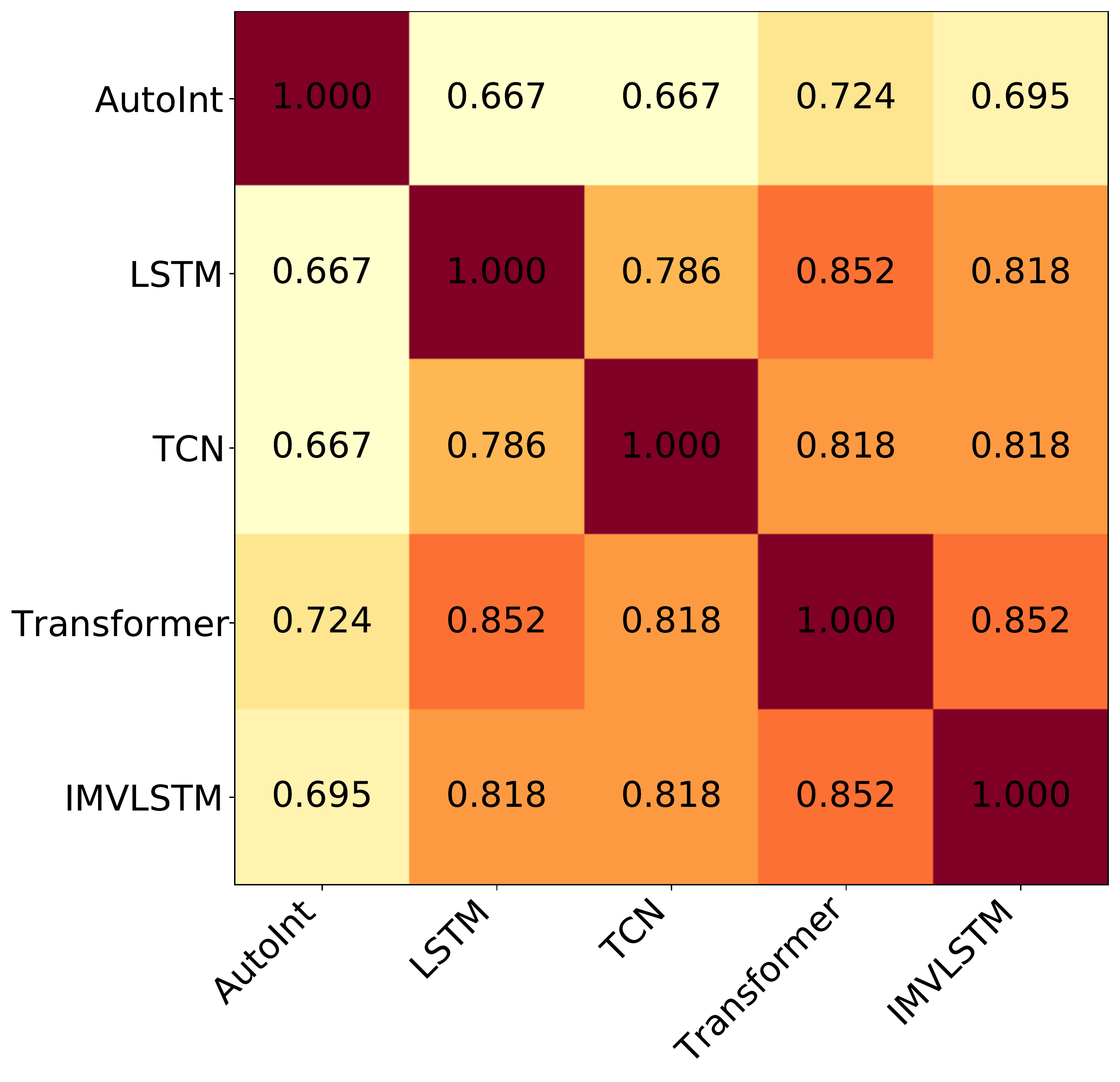}
    \caption{Jaccard similarity of top-50 most important features identified in all models.}
    \label{fig:jaccard-sim}
    \vspace{-2.5em}
\end{wrapfigure}

We further investigate and compare important features given by different prediction models with the best performing interpretablity method ArchDetect in Section~\ref{subsubsec:eval-of-interp} for a qualitative evaluation of its effectiveness. Since ArchDetect gives local feature importance for each data sample respectively, we aggregate local results for a global qualitative evaluation with following steps: (1) for each sample, get the rank of importance for each individual feature; (2) calculate the average of ranks for each feature over all data samples; (3) sort the averaged ranks of features from (2) as the global ordering of importance for all features. We then verify the effectiveness of feature importance estimation given by ArchDetect from following aspects:

\paragraph{Similarity of Important Features from Different Models}
\fref{fig:jaccard-sim} shows the Jaccard similarity of top-50 most important features identified in models. We observe that (1) the Jaccard similarity of top-50 most important features from any pair of two models is above 0.667; (2) each pair of models accepting sequential data (LSTM, TCN, Transformer, and IMVLSTM) has a Jaccard similarity over 0.786. This result demonstrates that ArchDetect identifies similar sets of important features when applied to various models, which is necessary for its correctness since the ground truth set of important features is unique.

\begin{figure}[htbp]
    \centering
    \includegraphics[width=\textwidth]{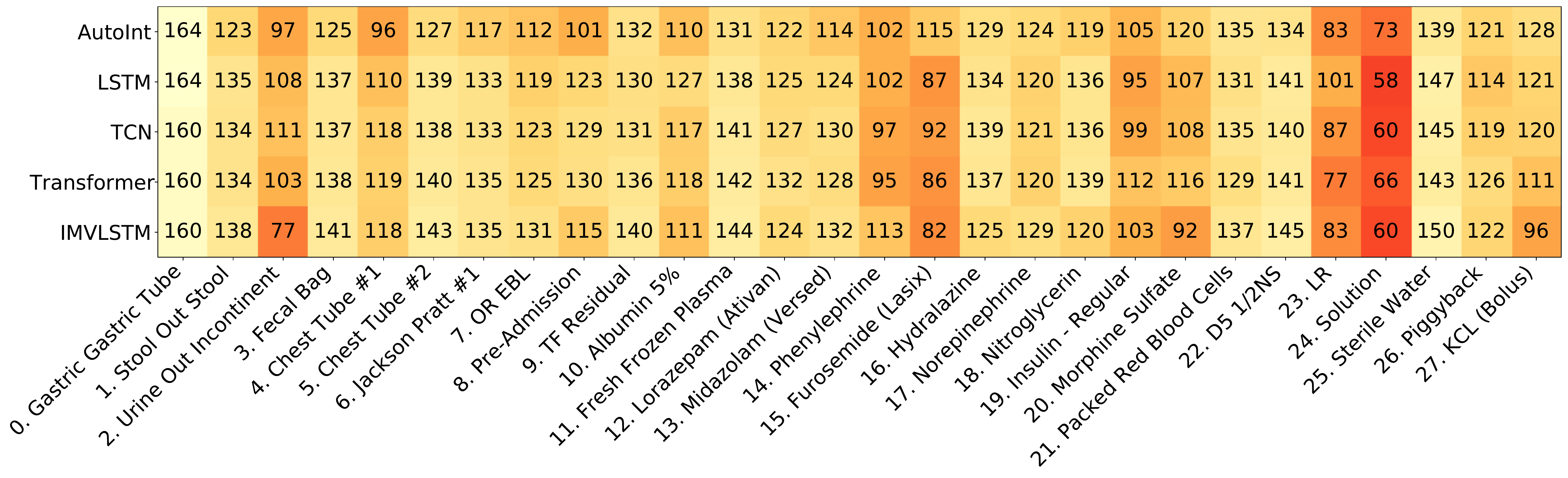}
    \caption{Visualization of Global Feature Importance Ranks (Feature 0-27). Each row is one prediction model and each column is one feature. The number at the $i$-th row and the $j$-th column represents the global feature importance rank of the $j$-th feature in the $i$-th model. Darker color means the feature is more important in the model.}
    \label{fig:vis-global-fea-imp-0}
\end{figure}

\begin{figure}[htbp]
    \centering
    \includegraphics[width=\textwidth]{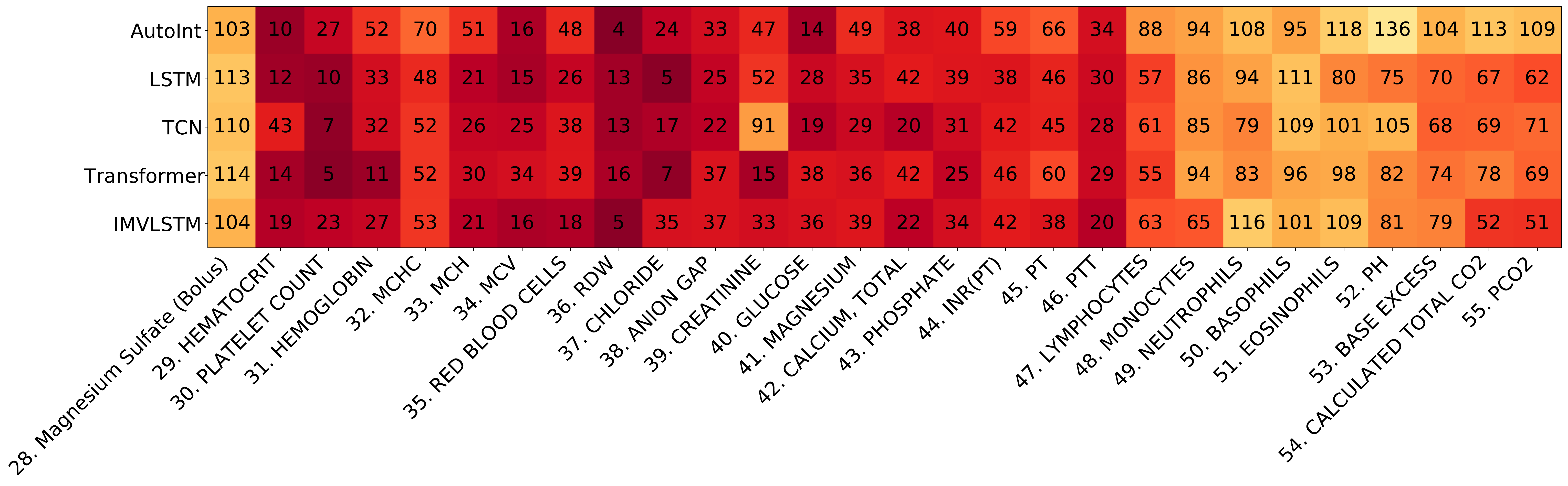}
    \caption{Visualization of Global Feature Importance Ranks (Feature 28-55).}
    \label{fig:vis-global-fea-imp-1}
\end{figure}

\begin{figure}[htbp]
    \centering
    \includegraphics[width=\textwidth]{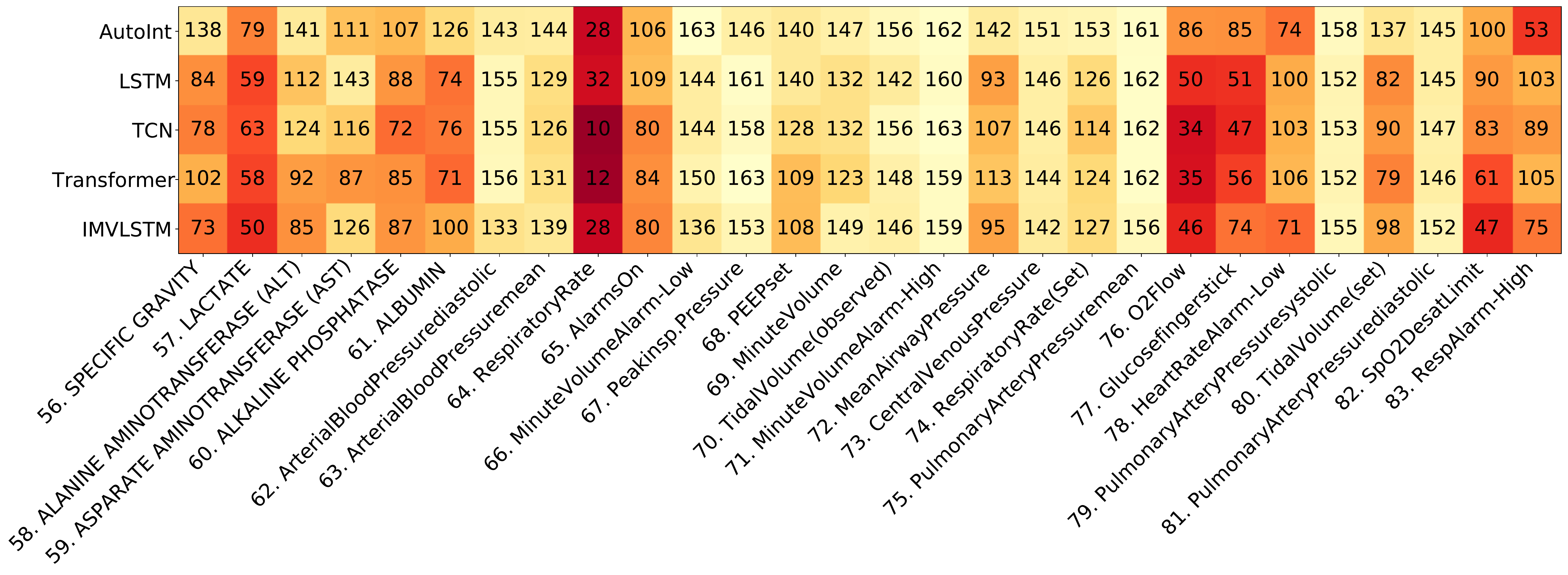}
    \caption{Visualization of Global Feature Importance Ranks (Feature 56-83).}
    \label{fig:vis-global-fea-imp-2}
\end{figure}

\begin{figure}[htbp]
    \centering
    \includegraphics[width=\textwidth]{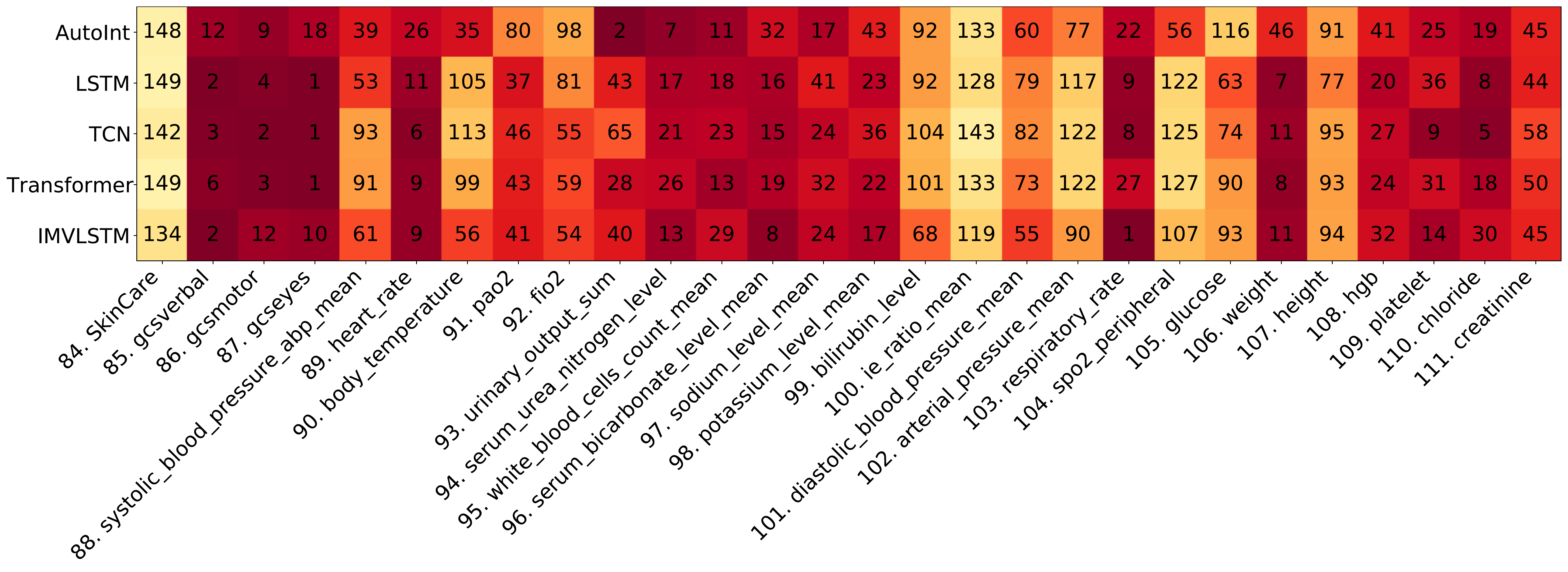}
    \caption{Visualization of Global Feature Importance Ranks (Feature 84-111).}
    \label{fig:vis-global-fea-imp-3}
\end{figure}

\begin{figure}[htbp]
    \centering
    \includegraphics[width=\textwidth]{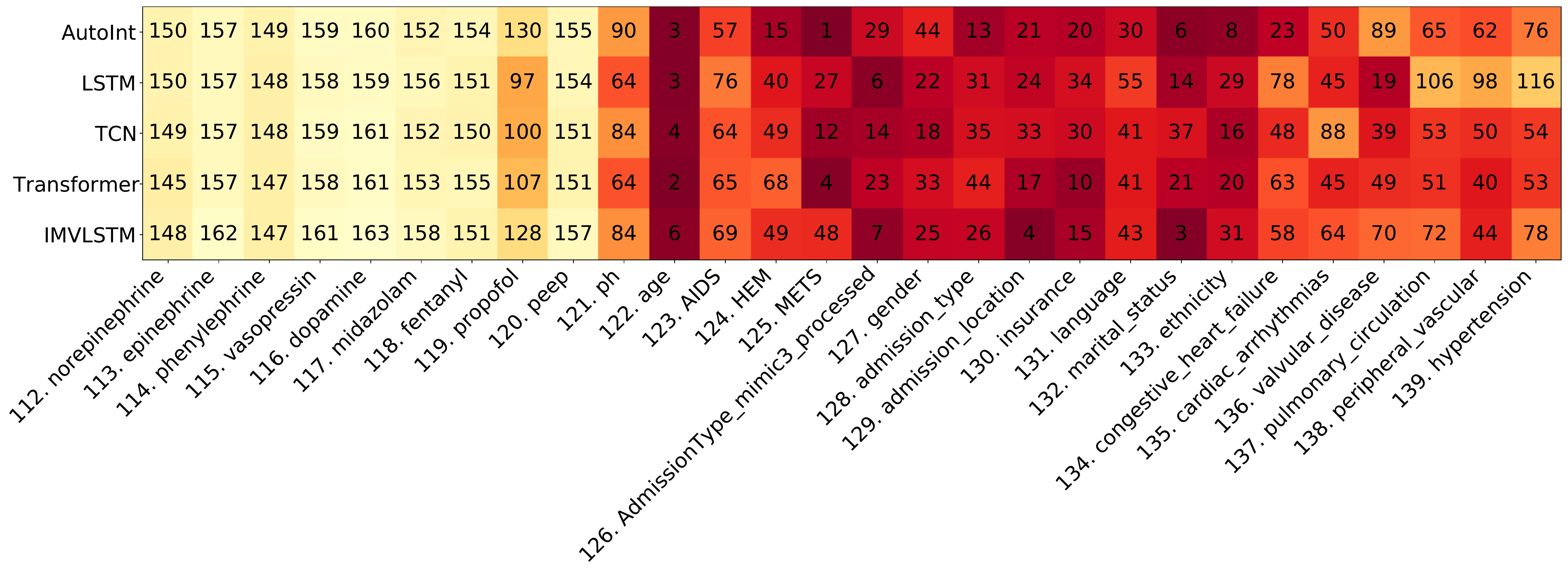}
    \caption{Visualization of Global Feature Importance Ranks (Feature 112-139).}
    \label{fig:vis-global-fea-imp-4}
\end{figure}

\begin{figure}[htbp]
    \centering
    \includegraphics[width=\textwidth]{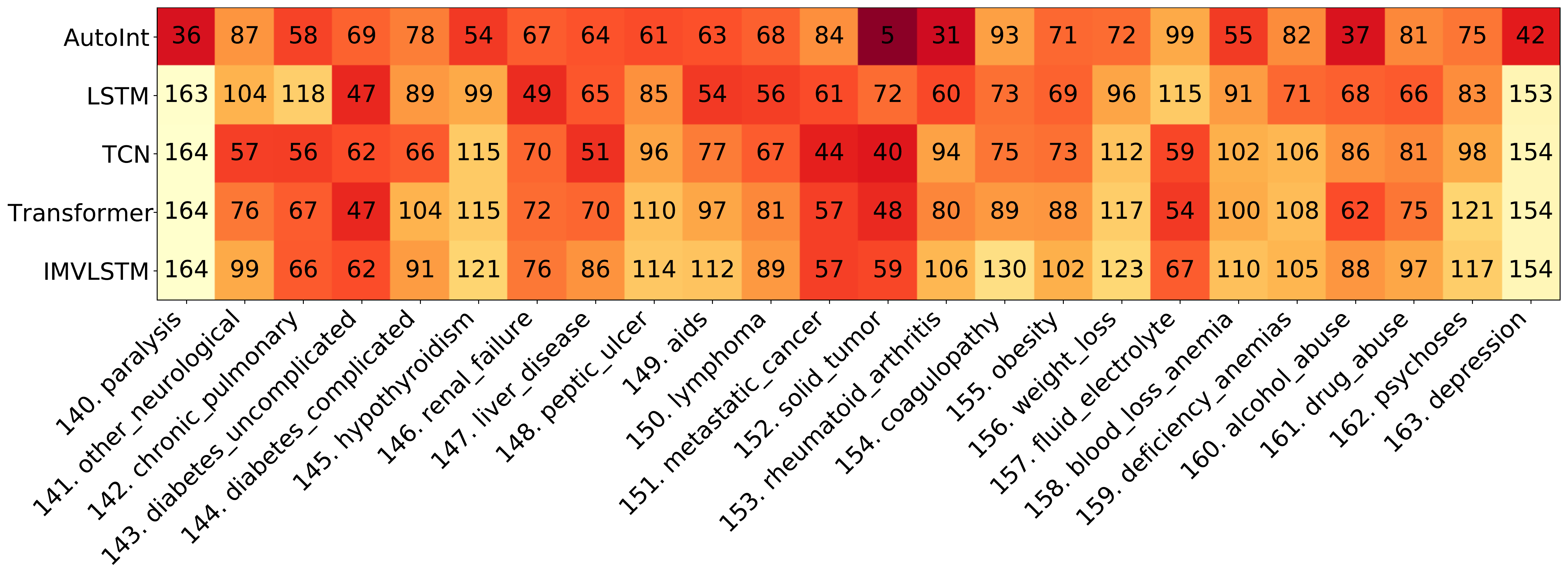}
    \caption{Visualization of Global Feature Importance Ranks (Feature 140-163).}
    \label{fig:vis-global-fea-imp-5}
\end{figure}

\paragraph{Visualization of Global Feature Importance Ranks}
\fref{fig:vis-global-fea-imp-0} -~\fref{fig:vis-global-fea-imp-5} visualize the aggregated global feature importance ranks from different models to (1) further demonstrate the similarity of feature importance estimation for different prediction models, and (2) to give an intuitive explanation of what features are important for mortality prediction tasks. We observe that there are several regions of features that all prediction models treat as important features: (1) \textbf{labevent features}, including feature 29-46 and feature 108-111, which contains laboratory based measurements of fluid of the patient's body; (2) \textbf{respiratory-related features}, including feature 64 (Respiratory Rate), feature 76 (O2Flow) and feature 103 (respiratory\_rate); (3) \textbf{SAPS-II features}, including feature 85-98, which are used in the Simplified Acute Physiology Score - II severity classification system; (4) \textbf{demographic features}, including feature 122 (age), feature 127 (gender), feature 130 (insurance), feature 132 (marital\_status), and feature 133 (ethnicity).

We find that the feature importance estimation results successfully identify critical features that are also used in the domain knowledge based SAPS-II system for severity classification, indicating the correctness of results. We also notice that demographic features play important roles in prediction models, which may raise the concern of fairness. We further investigate the fairness of data and models in the following section.


\section{Fairness Evaluation}
In this section, we first describe the set of demographic features considered as protected attributes. We then investigate the extent of which disparate treatment exists within the MIMIC-IV dataset.
Given that the in-hospital mortality predictors can be further developed and utilized in a down-stream decision-making policy, 
we further audit their performance in terms of fairness across various protected attributes.

\subsection{Protected Attributes}
MIMIC-IV came with a set of demographic features that are helpful for the task of auditing in-hospital mortality predictors for prediction fairness. Protected classes under the Equal Credit Opportunity Act (ECOA) include the following: \textit{age, color, marital status, national origin, race, recipient of public assistance, religion, sex}~\cite{chen2019fairness}. For our task, we consider a subset of such protected classes available within the dataset. To remove uncertainty within our analysis, we further identify and drop examples with unclear attributes, such as `None', `Unknown', or `Unable to obtain'. Table \ref{tab:attributes} lists the attributes and subgroups used within our analysis. Note that \textit{age} is grouped by quartiles. For a more in-depth look at the distribution of each subgroup, please refer to Table \ref{tab:dist-mor} in the Appendix section.
\vspace{-0.2cm}
\begin{table}[htbp]
\centering
\caption{Protected attributes and subgroups within MIMIC-IV.}
\label{tab:attributes}
\resizebox{0.87\textwidth}{!}{
\begin{tabular}{ll} \toprule
Protected Attributes & Groups \\ \midrule
Ethnicity         &  \texttt{[`ASIAN', `BLACK/AFRICAN AMERICAN', `HISPANIC/LATINO', `OTHER', `WHITE']}\\
Gender            &  \texttt{[`FEMALE', `MALE']}\\
Marital Status    &  \texttt{[`MARRIED', `SINGLE', `DIVORCED/WIDOWED']}\\
Age               & \texttt{[`<55 YRS', `55-67 YRS', `67-78 YRS', `>=78 YRS']} \\
Insurance         & \texttt{[`MEDICAID/MEDICARE', `PRIVATE']} \\\bottomrule
\end{tabular}}
\vspace{-0.3cm}
\end{table}

\subsection{Fair Treatment Analysis}
Disparate treatment is unlawful discrimination in US labor law. Title VII of the United States Civil Rights Act is created to prevent unequal treatment or behavior toward someone because of a protected attribute (e.g. race, gender, or religious beliefs). 
Although the type and duration of treatment received by patients are determined by multiple factors, analyzing treatment disparities in MIMIC-IV can give us insights in potential biases in treatment received by different groups.
Previously, there have been a few works pointing out the racial disparities in end-of-life care between cohorts of black and white patients within MIMIC-III \cite{Yarnell2017AssociationBI, LEE20169}. In a similar spirit, we additionally investigate treatment adoptions and duration across not only ethnicity, but also gender, age, marital status, and insurance type.

\subsubsection{Evaluation Method}
In MIMIC-IV, 5 categories of mechanical ventilation received by patients have been recorded: HighFlow, InvasiveVent, NonInvasiveVent, Oxygen, and Trach. We first extract the treatment duration and then label the patients with no record as no intervention adoption. If a patient had multiple spans, such as an intubation-extubation-reintubation, then we consider the patient’s treatment duration to be the sum of the individual spans.

\subsubsection{Results}
\begin{figure}
    \centering
    \begin{subfigure}[b]{\textwidth}
        \centering
        \caption{Average Intervention Adoption}\label{fig:fairness-interventions-adoption}\includegraphics[width=\textwidth]{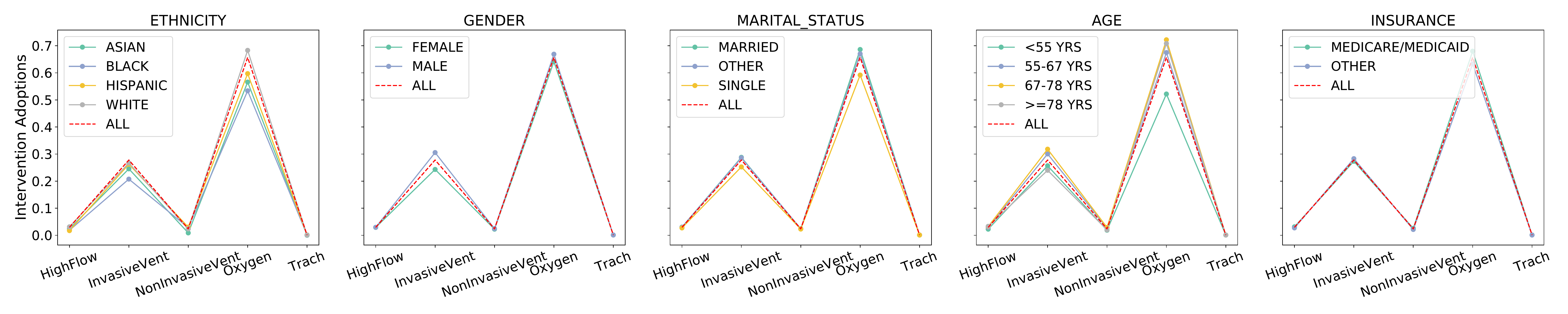}
    \end{subfigure}
    \hfill
    \begin{subfigure}[b]{\textwidth}
        \centering
        \caption{Average Intervention Hours}\label{fig:fairness-interventions-hours}\includegraphics[width=\textwidth]{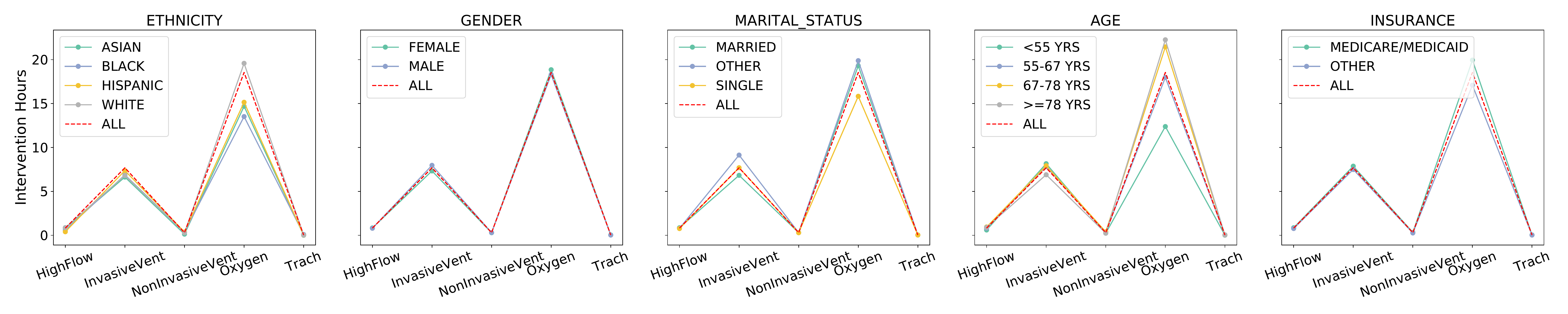}
    \end{subfigure}
    \caption{Average adoption and hours of intervention in general and in subjects from different groups.}
    \label{fig:fairness-dist-interventions}
\end{figure}

Figure \ref{fig:fairness-dist-interventions} plots the intervention adoption rate and intervention duration across different protected attributes. We observe that: (1) \textbf{There exists disparate treatments, which is most evident across different ethnic groups.} The first column in Figure \ref{fig:fairness-dist-interventions} indicates that on average Black and Hispanic cohorts are less likely to receive ventilation treatments, while also receiving a shorter treatment duration. Similarly, this is also observed across groups split by the marital status, where single patients tend to receive shorter and fewer ventilation treatments as opposed to married patients, and similarly with patients with public or private insurances. (2) \textbf{There are numerous hidden confounders in analyzing disparate treatment}. The fourth column in Figure \ref{fig:fairness-dist-interventions} indicates more treatments provided to older patients. However, one can imagine that cause of this is medically relevant as the older cohort tends to require more care. Similarly, patients with generous public insurance can more easily afford more treatments. In particular, we note that it is difficult to precisely determine whether the differences in treatment are due to intentional discrimination or to differences caused by other confounders. At the current junction, we suspect a close look at causal analysis can help address this problem.

\subsection{Fair Prediction Analysis}
Fairness in machine learning is a rapidly developing field with numerous definitions and metrics for prediction fairness with respect to two notions: individual and group fairness. For our binary classification task of in-hospital mortality prediction, we consider the group notion where a small number of protected demographic groups $G$ (such as racial groups) is fixed, and we then ask for the classification parity of certain statistics across all of these protected groups.

\subsubsection{Fairness Metrics}
Most recently, a multitude of statistical measures have been introduced for group fairness, most notable are statistics that ask for the equality of the false positive or negative rates across all groups $G$ (often known as \textit{`equal opportunity'} \cite{hardt2016equality}) or the equality of classification rates (also known as \textit{statistical parity}). Interestingly, it has been proven that some of the competing definitions and statistics previously proposed are mutually exclusive \cite{kleinberg2016inherent}. Thus, it is impossible to satisfy all of these fairness constraints.

In our case, it is often necessary for mortality assessment algorithms to explicitly consider health-related protected characteristics, especially the age of the patients. Hence, an age-neutral assessment score can systematically overestimate a young person's mortality risk, and can in turn encourage unnecessarily medical interventions. Similarly, enforcing equality of mortality classification rates can likewise lead to discriminatory decision making.

Hence, we choose AUC (area under the ROC curve) as our evaluation metrics to audit fairness across subgroups. First, it encompasses both FPR and FNR, which touches on the notion of equalized opportunity and equalized odds. Second, it is robust to class imbalance, which is especially important in the task of mortality prediction where mortality rates are $\sim 7\%$, Lastly, AUC is threshold agnostic, which does not necessitate setting a specific threshold for binary prediction that is used across all groups.

\subsubsection{Evaluation Method}
To evaluate fairness for each model on the MIMIC-IV dataset, we stratify the test set by groups (Table \ref{tab:attributes}), and compute the AUC for each protected group, similarly to \cite{lahoti2020fairness}. In addition, we also added a stratification for the patient group with the largest common comorbidity, with HEM/METS for patients with lymphoma, leukemia, multiple myeloma, and metastatic cancer. We report (1) AUC(min): minimum AUC over all protected groups, (2) AUC(macro-avg): macro-average over all protected group AUCs and (3) AUC(minority): AUC reported for the smallest protected group in the dataset. Higher AUC is better for all three metrics.

Additionally, as MIMIC-IV is an ongoing data collection effort, we also investigate the relationships between the predictive performance of the mortality predictors and the data distribution with respect to each protected group. It was shown in \cite{corbett2018measure} that if the risk distributions of protected groups in general differ, such as mortality rates, threshold-based decisions will typically yield error metrics that also differ by group. Hence, we are interested in studying the potential source of the bias/differences in predictive performances from the MIMIC-IV training set.

\subsubsection{Results}
\begin{figure}
    \centering
    \begin{subfigure}[b]{\textwidth}
        \centering
        \includegraphics[width=\textwidth]{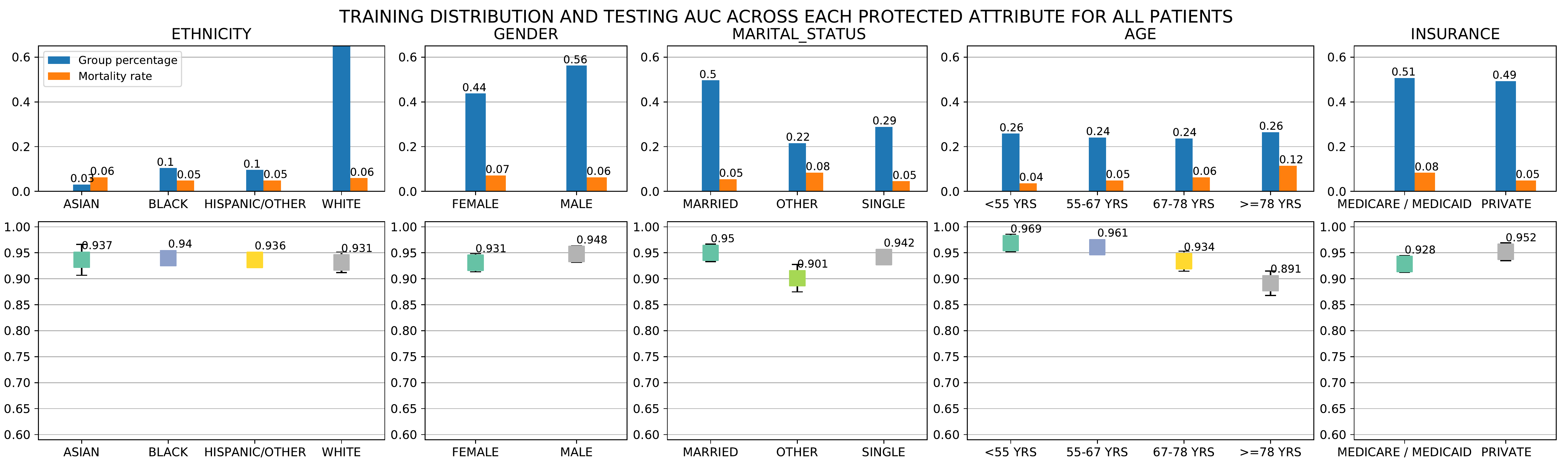}
    \end{subfigure}
    \hfill
    \begin{subfigure}[b]{\textwidth}
        \centering
        \includegraphics[width=\textwidth]{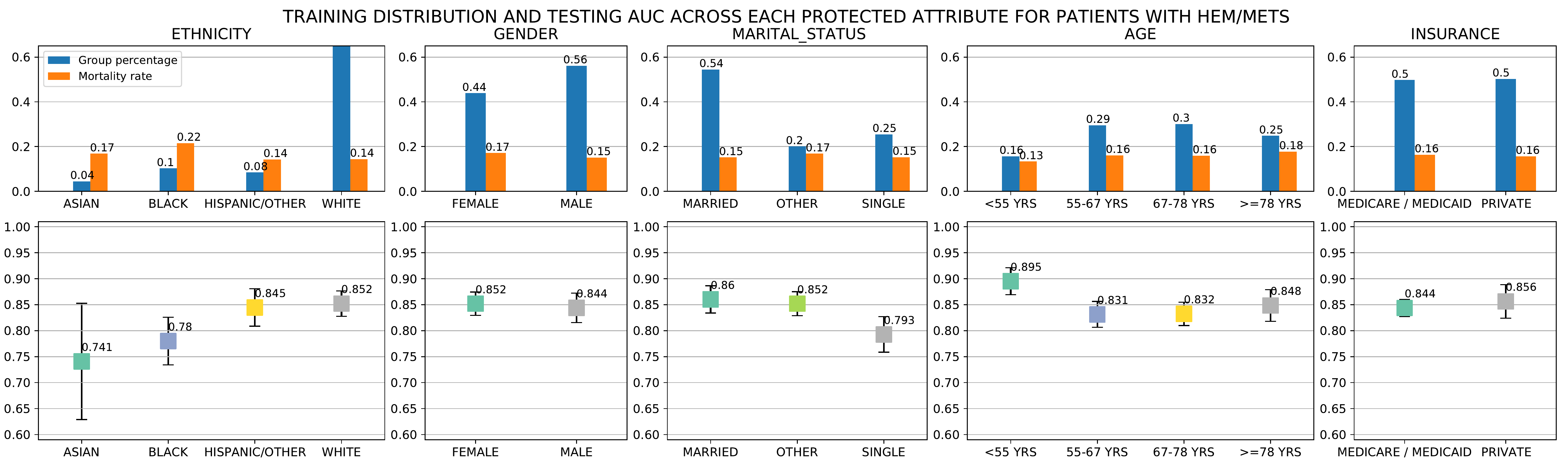}
    \end{subfigure}
    \caption{Training data distribution, mortality rates, and testing AUCs across each protected attribute for all patients and patients with HEM/METS, summarized over all five classifiers: AutoInt, LSTM, IMV-LSTM, TCN, and Transformer.}
    \label{fig:fairness-dist-auc}
    \vspace{-0.5cm}
\end{figure}

Figure \ref{fig:fairness-dist-auc} shows the training data distribution, mortality rates, and testing AUCs across each protected attribute for all patients and patients with HEM/METS, summarized over all five classifiers: AutoInt, LSTM, IMV-LSTM, TCN, and Transformer. Smaller gaps in AUC indicate equality in predictive performances, and larger gaps indicate potential inequalities. Table \ref{tab:fairness-auc} gives the quantitative results of the area under the curve (AUC). Higher values of AUCs for each of the min, avg, and minority AUC metrics indicate better predictive performance with respect to the protected groups.

We have the following observations: (1) \textbf{IMV-LSTM performs the best overall on fairness measure with respect to AUC across different protected groups}. Quantitatively, from Table \ref{tab:fairness-auc}, it is clear that IMV-LSTM has the highest AUC for both overall samples and the subgroups. We see that the minimum AUC for the protected subgroups is highest among the methods considered in this work. This indicates a higher lower bound over all protected attributes. Moreover, the AUC gap for minimum over protected groups is much larger  than the next best model, Transformer, for the patient groups with HEM, and METS. (2) \textbf{The in-hospital mortality predictors are in general fair, but less so for the subgroup of patients with the comorbidity HEM/METS}. From Figure \ref{fig:fairness-dist-auc}, we observe that the maximum AUC gap across all attributes is at most 0.08, which is smaller than the maximum AUC gap for patients with HEM and METS at 0.11. The difference is more pronounced in the Ethnicity class, but can similarly be observed for other protected classes. In general, we note that all models are quite fair across ethnic groups, with small deviations in gender, and patient's insurance. Across both sets of patients, we see that all classifiers are in general more accurate for younger patients (<55years) versus older patients. (3) \textbf{There exists a strong correlation between mortality rates and AUCs for each of the protected attributes.} We observe that there is a strong correlation between group mortality rates and group AUC, with Pearson's r=-0.922 and a p-value < .00001. This shows that groups with higher mortality rates indicate lower AUC scores. From Figure \ref{fig:fairness-dist-auc}, we also observe that data with imbalanced representation between each subgroup does not impact predictive performance substantively.

\begin{table}[htbp]
    \centering
    \caption{Summarized Area under the curve (AUC) performance of the in-hospital mortality predictors evaluated on sets of protected groups. Higher AUC indicates better predictive performance.}
    \label{tab:fairness-auc}
    \resizebox{\textwidth}{!}{
        \begin{tabular}{cccccc} \toprule
Methods     & Patient Group                            & AUC     & Minimum AUC               & Macro-average AUC        & AUC                              \\
            &                                          & Overall & over all protected groups & over all protected group & for the smallest protected group \\ \midrule
AutoInt     & All                                      & 0.900   & 0.832                     & 0.897                    & 0.882                            \\
LSTM        &                                          & 0.941   & 0.896                     & 0.939                    & 0.932                            \\
TCN         &                                          & 0.937   & 0.883                     & 0.936                    & 0.948                            \\
Transformer &                                          & 0.941   & 0.898                     & 0.939                    & 0.953                            \\
IMV-LSTM    &                                          & \textbf{0.955}   & \textbf{0.918}                     & \textbf{0.954}                    & \textbf{0.968}                            \\ \midrule
AutoInt     & \multirow{5}{*}{HEM, METS} & 0.795   & 0.546                     & 0.783                    & 0.546                            \\
LSTM        &                                          & 0.842   & 0.726                     & 0.830                    & 0.777                            \\
TCN         &                                          & 0.832   & 0.696                     & 0.822                    & 0.696                            \\
Transformer &                                          & 0.839   & 0.778                     & 0.830                    & 0.823                            \\
IMV-LSTM    &                                          & \textbf{0.884}   & \textbf{0.845}                     & \textbf{0.879}                    & \textbf{0.862}  \\ \bottomrule                         
\end{tabular}
}
\end{table}

\section{Interactions between Interpretability and Fairness}
Fairness and interpretability are two critical pillars of the recent push for fairness, accountability, and transparency within deep learning. Overall, most interpretability works concern with explaining how the input features impact the final prediction, whether through feature importance or attributions, interactions, and knowledge distillation. Fairness on the other hand concerns with fairness metrics, optimization for fairness constraints, and the trade-off between accuracy and fairness. However, to the best of our knowledge, few work attempts to answer the question of how can interpretability help with fairness. What can we learn from our interpretability methods that would indicate either algorithmic bias or representation bias? In this section, we present concrete evidence to establish the initial connection between the two areas, but admittedly leave the fully investigation on the strength of this interaction for future work.

\subsection{Feature Importance Correlation with Fairness Metrics}
\begin{figure}[t!]
    \centering
    \begin{subfigure}[b]{\textwidth}
        \centering
        \caption{DeepLift}\label{fig:interactions_DeepLift}\includegraphics[width=\textwidth]{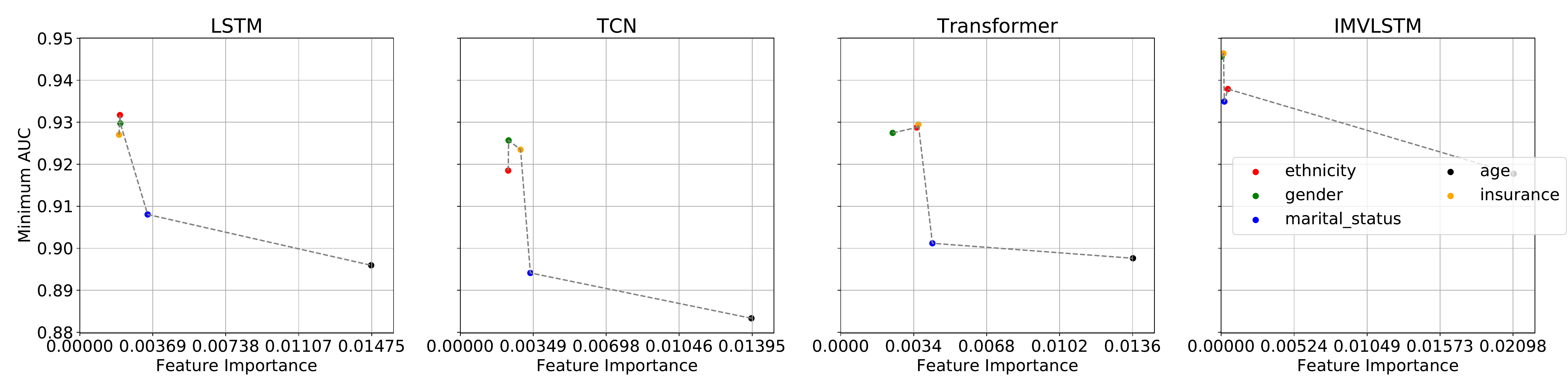}
    \end{subfigure}
    \hfill
    \begin{subfigure}[b]{\textwidth}
        \centering
        \caption{DeepLiftShap}\label{fig:interactions_DeepLiftShap}\includegraphics[width=\textwidth]{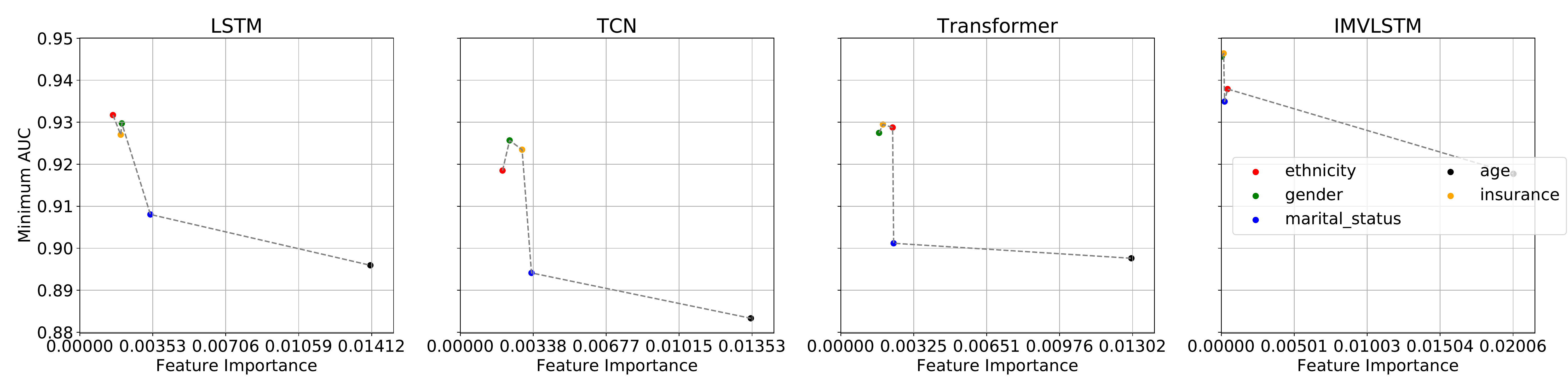}
    \end{subfigure}
    \caption{Interactions between \emph{Feature Importance} from two interpretability approaches and fairness evaluation value \emph{Min AUC} based on mortality predictions from four models.}
    \label{fig:interaction-plot}
    \vspace{-0.5cm}
\end{figure}

Given mortality predictions made by state-of-the-art models on MIMIC-IV, we study the connections between feature importance induced by different interpretation approaches and the fairness measures in Figure~\ref{fig:interaction-plot}. For all the five protected attributes, we compute their respective feature importance by averaging the values produced from interpretability models across time and patients. Taking the feature importance as $x$ axis and the minimum AUC from subgroups split by protected attributes as $y$ axis, we are expecting to see a decreasing trend, where more important features have a higher possibility to lead to performance divergence in the split subgroups. We observe the expected trend consistently among all prediction models, when the interpretability approach \emph{DeepLift} and \emph{DeepLiftShap} are utilized. As shown in Figure~\ref{fig:interaction-plot}, age (black dot) is the most important feature compared with other protected attributes and the accuracy difference between young and old is more obvious than other group divisions. Similarly, ethnicity (red dot) and gender (green dot) are the least important features, which leads to much higher minimum AUC than other protected attributes. We plotted but did not observe obvious connections between feature importance from other interpretability approaches and other two fairness evaluation metrics.
\subsection{Feature Importance Scores across Protected Attributes}
Interpretability often concerns with \textit{global} feature importance for the entire model and \textit{local} feature importance for an individual sample with respect to its prediction. Here, we consider the group feature importance that builds upon \textit{local} feature importance. Ideally, we want to measure how important each feature is across different groups with certain protected attributes. Hence, we define the group feature importance $g_i$ for feature $i$ and protected attribute $A$, $$ g_{i,A} = \frac{1}{N_A}\sum_{j=1}^{N_A} \phi^j_i,$$ where $N_A$ is the size of the group with attribute $A$, and $\phi^j_i$ is the $local$ feature importance of the feature $i$ for a person $j$ with attribute $A$. The parity between $g_{i,A}$ would indicate a parity in how each feature is being used for different groups within a certain class of protected attributes. In the MIMIC-IV setting, we are interested in the importance of each of the demographic features used for the in-mortality prediction across the protected subgroups.

\begin{figure}
    \centering
    \includegraphics[width=\textwidth]{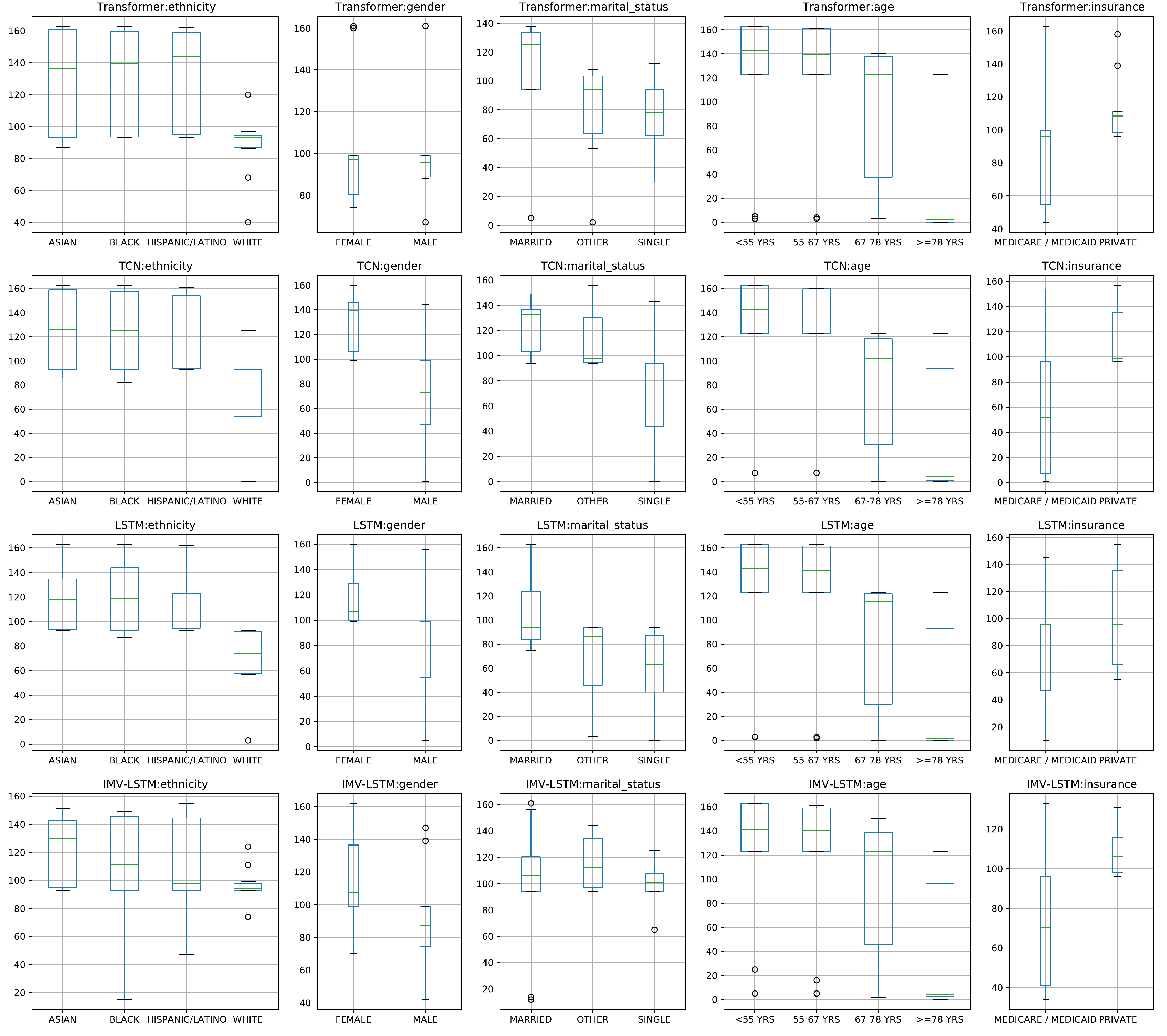}
    \caption{Feature rankings for each demographic feature for the four models: Transformer, TCN, LSTM, and IMV-LSTM, and each of the 12 interpretability methods: ArchDetect, DeepLiftShap, FeaturePermutation, IntegratedGradients, SaliencyNoiseTunnel, DeepLift, FeatureAblation, GradientShap, Occlusion, Saliency, and ShapleySampling.}
    \label{fig:group_f_import}
\end{figure}

Since the scales of the feature importance scores are different for each of the interpretability method, we calculate the group feature importance for each demographic feature and rank their importance relative to other features within each interpretability method. Additionally, since feature importance is provided for \{each hour timestep\} x \{each feature\} within the first 24 hours in the ICU, for all models, we additionally average the feature importance across timesteps.
Figure \ref{fig:group_f_import} presents the box plot of the feature rankings for each demographic feature for the four models: Transformer, TCN, LSTM, and IMV-LSTM, and each of the 12 interpretability methods: ArchDetect, DeepLiftShap, FeaturePermutation, IntegratedGradients, SaliencyNoiseTunnel, DeepLift, FeatureAblation, GradientShap, Occlusion, Saliency, and ShapleySampling. A lower ranking indicates higher feature importance.

We observe that similar trends exist across different models of varying architectures, where a demographic feature is more important (has lower ranking) for specific groups. Out of 164 features used for each timestep, the feature \texttt{ethnicity} has the highest feature importance for the \texttt{WHITE} patients, similarly for the \texttt{MALE} patients with the feature \texttt{gender}, and the age group \texttt{>= 78 YRS} with the feature \texttt{age}, and so on. The protected attribute age is the most intuitive in this setting, where in-hospital mortality predictors would attribute high importance to elderly patients since that is a strong signal for mortality prediction. A similar case can be made the feature insurance, as patients with Medicare are often elderly. However, it is less intuitive for the ethnicity feature, as to why one subgroup would use the ethnicity feature more strongly than the other subgroups. This stark parity exists for all models, even for different methods of interpretability to obtain feature importance. 

Again, it is difficult to identify the confounders or features that strongly correlate with the ethnicity feature, hence perhaps a causal perspective is most needed for this task. However, we do note that feature importance, especially when viewed as group importance, can concrete reveal how a feature is being used for different groups. Should this parity be considered as ``unfair'' and perhaps demographic features should not be used in training mortality classifiers. However, as we have also observed, age is such a strong indicator of mortality that omitting it would be detrimental to the classifiers' performance. We leave this connection for future work and emphasize the need to connect interpretability and fairness to explain what the model is doing and how can that explanation be used to reveal and/or audit potential bias within the model itself.

\section{Summary}
In this work, we conduct analysis on the MIMIC-IV dataset and several deep learning models in terms of model interpretability, dataset bias, algorithmic fairness, and the interaction between interpretability and fairness. We present quantitative evaluation of interpretability methods on deep learning models for mortality prediction, demonstrate the dataset bias in treatment in MIMIC-IV, verify the fairness of studied mortality prediction models, and reveal the disparities of feature importance among demographic subgroups. We will conduct further analysis from a causal perspective on the relation between the difference of feature importance and the difference of model outcomes among subgroups.
\clearpage
\bibliographystyle{unsrt}
\bibliography{references}

\newpage
\appendix
\setcounter{table}{0}
\renewcommand{\thetable}{A\arabic{table}}
\setcounter{figure}{0}
\renewcommand{\thefigure}{A\arabic{figure}}
\setcounter{algorithm}{0}
\renewcommand{\thealgorithm}{A\arabic{algorithm}}
\setcounter{equation}{0}
\renewcommand{\theequation}{A\arabic{equation}}
\section{Appendix}

\subsection{List of Features}

{\small
    \begin{longtable}[c]{llll}
        \caption{Full list of selected 164 features.\label{tab:full-list}}                                              \\
        \toprule
        $i$-th feature & Feature name                         & Group       & Tablename                                 \\
        \midrule
        0              & Gastric Gastric Tube                 & EHR         & mimic\_icu.outputevents                   \\
        1              & Stool Out Stool                      & EHR         & mimic\_icu.outputevents                   \\
        2              & Urine Out Incontinent                & EHR         & mimic\_icu.outputevents                   \\
        3              & Fecal Bag                            & EHR         & mimic\_icu.outputevents                   \\
        4              & Chest Tube \#1                       & EHR         & mimic\_icu.outputevents                   \\
        5              & Chest Tube \#2                       & EHR         & mimic\_icu.outputevents                   \\
        6              & Jackson Pratt \#1                    & EHR         & mimic\_icu.outputevents                   \\
        7              & OR EBL                               & EHR         & mimic\_icu.outputevents                   \\
        8              & Pre-Admission                        & EHR         & mimic\_icu.outputevents                   \\
        9              & TF Residual                          & EHR         & mimic\_icu.outputevents                   \\
        10             & Albumin 5\%                          & EHR         & mimic\_icu.inputevents                    \\
        11             & Fresh Frozen Plasma                  & EHR         & mimic\_icu.inputevents                    \\
        12             & Lorazepam (Ativan)                   & EHR         & mimic\_icu.inputevents                    \\
        13             & Midazolam (Versed)                   & EHR         & mimic\_icu.inputevents                    \\
        14             & Phenylephrine                        & EHR         & mimic\_icu.inputevents                    \\
        15             & Furosemide (Lasix)                   & EHR         & mimic\_icu.inputevents                    \\
        16             & Hydralazine                          & EHR         & mimic\_icu.inputevents                    \\
        17             & Norepinephrine                       & EHR         & mimic\_icu.inputevents                    \\
        18             & Nitroglycerin                        & EHR         & mimic\_icu.inputevents                    \\
        19             & Insulin - Regular                    & EHR         & mimic\_icu.inputevents                    \\
        20             & Morphine Sulfate                     & EHR         & mimic\_icu.inputevents                    \\
        21             & Packed Red Blood Cells               & EHR         & mimic\_icu.inputevents                    \\
        22             & D5 1/2NS                             & EHR         & mimic\_icu.inputevents                    \\
        23             & LR                                   & EHR         & mimic\_icu.inputevents                    \\
        24             & Solution                             & EHR         & mimic\_icu.inputevents                    \\
        25             & Sterile Water                        & EHR         & mimic\_icu.inputevents                    \\
        26             & Piggyback                            & EHR         & mimic\_icu.inputevents                    \\
        27             & KCL (Bolus)                          & EHR         & mimic\_icu.inputevents                    \\
        28             & Magnesium Sulfate (Bolus)            & EHR         & mimic\_icu.inputevents                    \\
        29             & HEMATOCRIT                           & EHR         & mimic\_hosp.labevents                     \\
        30             & PLATELET COUNT                       & EHR         & mimic\_hosp.labevents                     \\
        31             & HEMOGLOBIN                           & EHR         & mimic\_hosp.labevents                     \\
        32             & MCHC                                 & EHR         & mimic\_hosp.labevents                     \\
        33             & MCH                                  & EHR         & mimic\_hosp.labevents                     \\
        34             & MCV                                  & EHR         & mimic\_hosp.labevents                     \\
        35             & RED BLOOD CELLS                      & EHR         & mimic\_hosp.labevents                     \\
        36             & RDW                                  & EHR         & mimic\_hosp.labevents                     \\
        37             & CHLORIDE                             & EHR         & mimic\_hosp.labevents                     \\
        38             & ANION GAP                            & EHR         & mimic\_hosp.labevents                     \\
        39             & CREATININE                           & EHR         & mimic\_hosp.labevents                     \\
        40             & GLUCOSE                              & EHR         & mimic\_hosp.labevents                     \\
        41             & MAGNESIUM                            & EHR         & mimic\_hosp.labevents                     \\
        42             & CALCIUM, TOTAL                       & EHR         & mimic\_hosp.labevents                     \\
        43             & PHOSPHATE                            & EHR         & mimic\_hosp.labevents                     \\
        44             & INR(PT)                              & EHR         & mimic\_hosp.labevents                     \\
        45             & PT                                   & EHR         & mimic\_hosp.labevents                     \\
        46             & PTT                                  & EHR         & mimic\_hosp.labevents                     \\
        47             & LYMPHOCYTES                          & EHR         & mimic\_hosp.labevents                     \\
        48             & MONOCYTES                            & EHR         & mimic\_hosp.labevents                     \\
        49             & NEUTROPHILS                          & EHR         & mimic\_hosp.labevents                     \\
        50             & BASOPHILS                            & EHR         & mimic\_hosp.labevents                     \\
        51             & EOSINOPHILS                          & EHR         & mimic\_hosp.labevents                     \\
        52             & PH                                   & EHR         & mimic\_hosp.labevents                     \\
        53             & BASE EXCESS                          & EHR         & mimic\_hosp.labevents                     \\
        54             & CALCULATED TOTAL CO2                 & EHR         & mimic\_hosp.labevents                     \\
        55             & PCO2                                 & EHR         & mimic\_hosp.labevents                     \\
        56             & SPECIFIC GRAVITY                     & EHR         & mimic\_hosp.labevents                     \\
        \hline
        57             & LACTATE                              & EHR         & mimic\_hosp.labevents                     \\
        58             & ALANINE AMINOTRANSFERASE (ALT)       & EHR         & mimic\_hosp.labevents                     \\
        59             & ASPARATE AMINOTRANSFERASE (AST)      & EHR         & mimic\_hosp.labevents                     \\
        60             & ALKALINE PHOSPHATASE                 & EHR         & mimic\_hosp.labevents                     \\
        61             & ALBUMIN                              & EHR         & mimic\_hosp.labevents                     \\
        62             & ArterialBloodPressurediastolic       & EHR         & mimic\_icu.chartevents                    \\
        63             & ArterialBloodPressuremean            & EHR         & mimic\_icu.chartevents                    \\
        64             & RespiratoryRate                      & EHR         & mimic\_icu.chartevents                    \\
        65             & AlarmsOn                             & EHR         & mimic\_icu.chartevents                    \\
        66             & MinuteVolumeAlarm-Low                & EHR         & mimic\_icu.chartevents                    \\
        67             & Peakinsp.Pressure                    & EHR         & mimic\_icu.chartevents                    \\
        68             & PEEPset                              & EHR         & mimic\_icu.chartevents                    \\
        69             & MinuteVolume                         & EHR         & mimic\_icu.chartevents                    \\
        70             & TidalVolume(observed)                & EHR         & mimic\_icu.chartevents                    \\
        71             & MinuteVolumeAlarm-High               & EHR         & mimic\_icu.chartevents                    \\
        72             & MeanAirwayPressure                   & EHR         & mimic\_icu.chartevents                    \\
        73             & CentralVenousPressure                & EHR         & mimic\_icu.chartevents                    \\
        74             & RespiratoryRate(Set)                 & EHR         & mimic\_icu.chartevents                    \\
        75             & PulmonaryArteryPressuremean          & EHR         & mimic\_icu.chartevents                    \\
        76             & O2Flow                               & EHR         & mimic\_icu.chartevents                    \\
        77             & Glucosefingerstick                   & EHR         & mimic\_icu.chartevents                    \\
        78             & HeartRateAlarm-Low                   & EHR         & mimic\_icu.chartevents                    \\
        79             & PulmonaryArteryPressuresystolic      & EHR         & mimic\_icu.chartevents                    \\
        80             & TidalVolume(set)                     & EHR         & mimic\_icu.chartevents                    \\
        81             & PulmonaryArteryPressurediastolic     & EHR         & mimic\_icu.chartevents                    \\
        82             & SpO2DesatLimit                       & EHR         & mimic\_icu.chartevents                    \\
        83             & RespAlarm-High                       & EHR         & mimic\_icu.chartevents                    \\
        84             & SkinCare                             & EHR         & mimic\_icu.chartevents                    \\
        85             & gcsverbal                            & EHR         & mimic\_icu.chartevents                    \\
        86             & gcsmotor                             & EHR         & mimic\_icu.chartevents                    \\
        87             & gcseyes                              & EHR         & mimic\_icu.chartevents                    \\
        88             & systolic\_blood\_pressure\_abp\_mean & EHR         & mimic\_icu.chartevents                    \\
        89             & heart\_rate                          & EHR         & mimic\_icu.chartevents                    \\
        90             & body\_temperature                    & EHR         & mimic\_icu.chartevents                    \\
        91             & pao2                                 & EHR         & mimic\_hosp.labevents                     \\
        92             & fio2                                 & EHR         & mimic\_hosp.labevents                     \\
        93             & urinary\_output\_sum                 & EHR         & mimic\_icu.outputevents                   \\
        94             & serum\_urea\_nitrogen\_level         & EHR         & mimic\_hosp.labevents                     \\
        95             & white\_blood\_cells\_count\_mean     & EHR         & mimic\_hosp.labevents                     \\
        96             & serum\_bicarbonate\_level\_mean      & EHR         & mimic\_hosp.labevents                     \\
        97             & sodium\_level\_mean                  & EHR         & mimic\_hosp.labevents                     \\
        98             & potassium\_level\_mean               & EHR         & mimic\_hosp.labevents                     \\
        99             & bilirubin\_level                     & EHR         & mimic\_hosp.labevents                     \\
        100            & ie\_ratio\_mean                      & EHR         & mimic\_icu.chartevents                    \\
        101            & diastolic\_blood\_pressure\_mean     & EHR         & mimic\_icu.chartevents                    \\
        102            & arterial\_pressure\_mean             & EHR         & mimic\_icu.chartevents                    \\
        103            & respiratory\_rate                    & EHR         & mimic\_icu.chartevents                    \\
        104            & spo2\_peripheral                     & EHR         & mimic\_icu.chartevents                    \\
        105            & glucose                              & EHR         & mimic\_icu.chartevents                    \\
        106            & weight                               & EHR         & mimic\_icu.chartevents                    \\
        107            & height                               & EHR         & mimic\_icu.chartevents                    \\
        108            & hgb                                  & EHR         & mimic\_hosp.labevents                     \\
        109            & platelet                             & EHR         & mimic\_hosp.labevents                     \\
        110            & chloride                             & EHR         & mimic\_hosp.labevents                     \\
        111            & creatinine                           & EHR         & mimic\_hosp.labevents                     \\
        112            & norepinephrine                       & EHR         & mimic\_icu.chartevents                    \\
        113            & epinephrine                          & EHR         & mimic\_icu.chartevents                    \\
        114            & phenylephrine                        & EHR         & mimic\_icu.chartevents                    \\
        115            & vasopressin                          & EHR         & mimic\_icu.chartevents                    \\
        116            & dopamine                             & EHR         & mimic\_icu.chartevents                    \\
        117            & midazolam                            & EHR         & mimic\_icu.chartevents                    \\
        118            & fentanyl                             & EHR         & mimic\_icu.chartevents                    \\
        119            & propofol                             & EHR         & mimic\_icu.chartevents                    \\
        120            & peep                                 & EHR         & mimic\_hosp.labevents                     \\
        \hline
        121            & ph                                   & EHR         & mimic\_hosp.labevents                     \\
        122            & age                                  & Demographic & mimic\_core.patients, mimic\_icu.icustays \\
        123            & AIDS                                 & Comorbidity & mimic\_hosp.diagnoses\_icd                \\
        124            & HEM                                  & Comorbidity & mimic\_hosp.diagnoses\_icd                \\
        125            & METS                                 & Comorbidity & mimic\_hosp.diagnoses\_icd                \\
        126            & AdmissionType\_mimic3\_processed     & Admission   & mimic\_core.admissions                    \\
        127            & gender                               & Demographic & mimic\_core.patients                      \\
        128            & admission\_type                      & Admission   & mimic\_core.admissions                    \\
        129            & admission\_location                  & Admission   & mimic\_core.admissions                    \\
        130            & insurance                            & Admission   & mimic\_core.admissions                    \\
        131            & language                             & Demographic & mimic\_core.admissions                    \\
        132            & marital\_status                      & Demographic & mimic\_core.admissions                    \\
        133            & ethnicity                            & Demographic & mimic\_core.admissions                    \\
        134            & congestive\_heart\_failure           & Comorbidity & mimic\_hosp.diagnoses\_icd                \\
        135            & cardiac\_arrhythmias                 & Comorbidity & mimic\_hosp.diagnoses\_icd                \\
        136            & valvular\_disease                    & Comorbidity & mimic\_hosp.diagnoses\_icd                \\
        137            & pulmonary\_circulation               & Comorbidity & mimic\_hosp.diagnoses\_icd                \\
        138            & peripheral\_vascular                 & Comorbidity & mimic\_hosp.diagnoses\_icd                \\
        139            & hypertension                         & Comorbidity & mimic\_hosp.diagnoses\_icd                \\
        140            & paralysis                            & Comorbidity & mimic\_hosp.diagnoses\_icd                \\
        141            & other\_neurological                  & Comorbidity & mimic\_hosp.diagnoses\_icd                \\
        142            & chronic\_pulmonary                   & Comorbidity & mimic\_hosp.diagnoses\_icd                \\
        143            & diabetes\_uncomplicated              & Comorbidity & mimic\_hosp.diagnoses\_icd                \\
        144            & diabetes\_complicated                & Comorbidity & mimic\_hosp.diagnoses\_icd                \\
        145            & hypothyroidism                       & Comorbidity & mimic\_hosp.diagnoses\_icd                \\
        146            & renal\_failure                       & Comorbidity & mimic\_hosp.diagnoses\_icd                \\
        147            & liver\_disease                       & Comorbidity & mimic\_hosp.diagnoses\_icd                \\
        148            & peptic\_ulcer                        & Comorbidity & mimic\_hosp.diagnoses\_icd                \\
        149            & aids                                 & Comorbidity & mimic\_hosp.diagnoses\_icd                \\
        150            & lymphoma                             & Comorbidity & mimic\_hosp.diagnoses\_icd                \\
        151            & metastatic\_cancer                   & Comorbidity & mimic\_hosp.diagnoses\_icd                \\
        152            & solid\_tumor                         & Comorbidity & mimic\_hosp.diagnoses\_icd                \\
        153            & rheumatoid\_arthritis                & Comorbidity & mimic\_hosp.diagnoses\_icd                \\
        154            & coagulopathy                         & Comorbidity & mimic\_hosp.diagnoses\_icd                \\
        155            & obesity                              & Comorbidity & mimic\_hosp.diagnoses\_icd                \\
        156            & weight\_loss                         & Comorbidity & mimic\_hosp.diagnoses\_icd                \\
        157            & fluid\_electrolyte                   & Comorbidity & mimic\_hosp.diagnoses\_icd                \\
        158            & blood\_loss\_anemia                  & Comorbidity & mimic\_hosp.diagnoses\_icd                \\
        159            & deficiency\_anemias                  & Comorbidity & mimic\_hosp.diagnoses\_icd                \\
        160            & alcohol\_abuse                       & Comorbidity & mimic\_hosp.diagnoses\_icd                \\
        161            & drug\_abuse                          & Comorbidity & mimic\_hosp.diagnoses\_icd                \\
        162            & psychoses                            & Comorbidity & mimic\_hosp.diagnoses\_icd                \\
        163            & depression                           & Comorbidity & mimic\_hosp.diagnoses\_icd                \\
        \bottomrule
    \end{longtable}
}

\subsection{Distributions of Demographic, Admission, and Comorbidity Features}

{\tiny \begin{longtable}{lllll}
        \caption{Distributions of demographic, admission and comorbidity features grouped by the in-hospital mortality label.\label{tab:dist-mor}}              \\
        \toprule
        Feature                          & Value Name                             & In-hospital Mortality = 0 & In-hospital Mortality = 1 & All                 \\
        \midrule
        age                              & [.25, .50, .75] quantile               & 54.51, 66.63, 77.94       & 62.14, 74.44, 84.20       & 55.02, 67.13, 78.59 \\
        AIDS                             & Negative                               & 99.57\% [39644]           & 99.50\% [3172]            & 99.56\% [42816]     \\
                                         & Positive                               & 0.43\% [173]              & 0.50\% [16]               & 0.44\% [189]        \\
        HEM                              & Negative                               & 97.67\% [38890]           & 94.76\% [3021]            & 97.46\% [41911]     \\
                                         & Positive                               & 2.33\% [927]              & 5.24\% [167]              & 2.54\% [1094]       \\
        METS                             & Negative                               & 94.81\% [37749]           & 86.89\% [2770]            & 94.22\% [40519]     \\
                                         & Positive                               & 5.19\% [2068]             & 13.11\% [418]             & 5.78\% [2486]       \\
        AdmissionType\_mimic3\_processed & medical                                & 68.17\% [27145]           & 83.28\% [2655]            & 69.29\% [29800]     \\
                                         & scheduled                              & 3.57\% [1420]             & 0.44\% [14]               & 3.33\% [1434]       \\
                                         & unscheduled                            & 28.26\% [11252]           & 16.28\% [519]             & 27.37\% [11771]     \\
        gender                           & F                                      & 43.81\% [17443]           & 47.15\% [1503]            & 44.06\% [18946]     \\
                                         & M                                      & 56.19\% [22374]           & 52.85\% [1685]            & 55.94\% [24059]     \\
        admission\_type                  & AMBULATORY OBSERVATION                 & 0.02\% [7]                & 0.00\% [0]                & 0.02\% [7]          \\
                                         & DIRECT EMER.                           & 3.56\% [1418]             & 3.23\% [103]              & 3.54\% [1521]       \\
                                         & DIRECT OBSERVATION                     & 0.09\% [37]               & 0.06\% [2]                & 0.09\% [39]         \\
                                         & ELECTIVE                               & 4.39\% [1749]             & 0.78\% [25]               & 4.13\% [1774]       \\
                                         & EU OBSERVATION                         & 0.22\% [86]               & 0.03\% [1]                & 0.20\% [87]         \\
                                         & EW EMER.                               & 50.37\% [20057]           & 61.48\% [1960]            & 51.20\% [22017]     \\

                                         & OBSERVATION ADMIT                      & 10.53\% [4192]            & 10.04\% [320]             & 10.49\% [4512]      \\
        \hline
                                         & SURGICAL SAME DAY ADMISSION            & 12.66\% [5040]            & 1.35\% [43]               & 11.82\% [5083]      \\
                                         & URGENT                                 & 18.16\% [7231]            & 23.02\% [734]             & 18.52\% [7965]      \\
        admission\_location              & AMBULATORY SURGERY TRANSFER            & 0.06\% [23]               & 0.03\% [1]                & 0.06\% [24]         \\
                                         & CLINIC REFERRAL                        & 0.83\% [330]              & 1.60\% [51]               & 0.89\% [381]        \\
                                         & EMERGENCY ROOM                         & 48.93\% [19483]           & 58.91\% [1878]            & 49.67\% [21361]     \\
                                         & INFORMATION NOT AVAILABLE              & 0.29\% [117]              & 0.41\% [13]               & 0.30\% [130]        \\
                                         & INTERNAL TRANSFER TO OR FROM PSYCH     & 0.01\% [3]                & 0.00\% [0]                & 0.01\% [3]          \\
                                         & PACU                                   & 0.56\% [222]              & 0.31\% [10]               & 0.54\% [232]        \\
                                         & PHYSICIAN REFERRAL                     & 24.22\% [9643]            & 7.59\% [242]              & 22.99\% [9885]      \\
                                         & PROCEDURE SITE                         & 1.69\% [673]              & 0.53\% [17]               & 1.60\% [690]        \\
                                         & TRANSFER FROM HOSPITAL                 & 21.20\% [8442]            & 27.51\% [877]             & 21.67\% [9319]      \\
                                         & TRANSFER FROM SKILLED NURSING FACILITY & 0.74\% [293]              & 1.54\% [49]               & 0.80\% [342]        \\
                                         & WALK-IN/SELF REFERRAL                  & 1.48\% [588]              & 1.57\% [50]               & 1.48\% [638]        \\
        insurance                        & Medicaid                               & 7.15\% [2846]             & 5.96\% [190]              & 7.06\% [3036]       \\
                                         & Medicare                               & 42.77\% [17029]           & 54.80\% [1747]            & 43.66\% [18776]     \\
                                         & Other                                  & 50.08\% [19942]           & 39.24\% [1251]            & 49.28\% [21193]     \\
        language                         & ?                                      & 9.78\% [3894]             & 10.92\% [348]             & 9.86\% [4242]       \\
                                         & ENGLISH                                & 90.22\% [35923]           & 89.08\% [2840]            & 90.14\% [38763]     \\
        marital\_status                  & DIVORCED                               & 7.17\% [2855]             & 5.58\% [178]              & 7.05\% [3033]       \\
                                         & MARRIED                                & 46.68\% [18588]           & 40.15\% [1280]            & 46.20\% [19868]     \\
                                         & None                                   & 6.63\% [2641]             & 17.57\% [560]             & 7.44\% [3201]       \\
                                         & SINGLE                                 & 27.08\% [10784]           & 19.64\% [626]             & 26.53\% [11410]     \\
                                         & WIDOWED                                & 12.43\% [4949]            & 17.06\% [544]             & 12.77\% [5493]      \\
        ethnicity                        & AMERICAN INDIAN/ALASKA NATIVE          & 0.17\% [66]               & 0.06\% [2]                & 0.16\% [68]         \\
                                         & ASIAN                                  & 2.88\% [1147]             & 3.26\% [104]              & 2.91\% [1251]       \\
                                         & BLACK/AFRICAN AMERICAN                 & 9.31\% [3708]             & 7.34\% [234]              & 9.17\% [3942]       \\
                                         & HISPANIC/LATINO                        & 3.60\% [1432]             & 2.51\% [80]               & 3.52\% [1512]       \\
                                         & OTHER                                  & 4.81\% [1914]             & 4.14\% [132]              & 4.76\% [2046]       \\
                                         & UNABLE TO OBTAIN                       & 1.33\% [531]              & 2.20\% [70]               & 1.40\% [601]        \\
                                         & UNKNOWN                                & 9.06\% [3609]             & 20.14\% [642]             & 9.88\% [4251]       \\
                                         & WHITE                                  & 68.84\% [27410]           & 60.35\% [1924]            & 68.21\% [29334]     \\
        congestive\_heart\_failure       & Negative                               & 98.66\% [39284]           & 98.90\% [3153]            & 98.68\% [42437]     \\
                                         & Positive                               & 1.34\% [533]              & 1.10\% [35]               & 1.32\% [568]        \\
        cardiac\_arrhythmias             & Negative                               & 99.56\% [39640]           & 99.97\% [3187]            & 99.59\% [42827]     \\
                                         & Positive                               & 0.44\% [177]              & 0.03\% [1]                & 0.41\% [178]        \\
        valvular\_disease                & Negative                               & 96.11\% [38267]           & 99.50\% [3172]            & 96.36\% [41439]     \\
                                         & Positive                               & 3.89\% [1550]             & 0.50\% [16]               & 3.64\% [1566]       \\
        pulmonary\_circulation           & Negative                               & 99.37\% [39568]           & 99.34\% [3167]            & 99.37\% [42735]     \\
                                         & Positive                               & 0.63\% [249]              & 0.66\% [21]               & 0.63\% [270]        \\
        peripheral\_vascular             & Negative                               & 98.78\% [39333]           & 98.78\% [3149]            & 98.78\% [42482]     \\
                                         & Positive                               & 1.22\% [484]              & 1.22\% [39]               & 1.22\% [523]        \\
        hypertension                     & Negative                               & 99.68\% [39689]           & 99.91\% [3185]            & 99.70\% [42874]     \\
                                         & Positive                               & 0.32\% [128]              & 0.09\% [3]                & 0.30\% [131]        \\
        paralysis                        & Negative                               & 100.00\% [39816]          & 99.97\% [3187]            & 100.00\% [43003]    \\
                                         & Positive                               & 0.00\% [1]                & 0.03\% [1]                & 0.00\% [2]          \\
        other\_neurological              & Negative                               & 99.40\% [39577]           & 99.72\% [3179]            & 99.42\% [42756]     \\
                                         & Positive                               & 0.60\% [240]              & 0.28\% [9]                & 0.58\% [249]        \\
        chronic\_pulmonary               & Negative                               & 99.52\% [39625]           & 99.78\% [3181]            & 99.54\% [42806]     \\
                                         & Positive                               & 0.48\% [192]              & 0.22\% [7]                & 0.46\% [199]        \\
        diabetes\_uncomplicated          & Negative                               & 99.40\% [39578]           & 99.97\% [3187]            & 99.44\% [42765]     \\
                                         & Positive                               & 0.60\% [239]              & 0.03\% [1]                & 0.56\% [240]        \\
        diabetes\_complicated            & Negative                               & 99.73\% [39710]           & 99.87\% [3184]            & 99.74\% [42894]     \\
                                         & Positive                               & 0.27\% [107]              & 0.13\% [4]                & 0.26\% [111]        \\
        hypothyroidism                   & Negative                               & 99.99\% [39813]           & 100.00\% [3188]           & 99.99\% [43001]     \\
                                         & Positive                               & 0.01\% [4]                & 0.00\% [0]                & 0.01\% [4]          \\
        renal\_failure                   & Negative                               & 99.88\% [39768]           & 99.97\% [3187]            & 99.88\% [42955]     \\
                                         & Positive                               & 0.12\% [49]               & 0.03\% [1]                & 0.12\% [50]         \\
        liver\_disease                   & Negative                               & 99.45\% [39597]           & 99.25\% [3164]            & 99.43\% [42761]     \\
                                         & Positive                               & 0.55\% [220]              & 0.75\% [24]               & 0.57\% [244]        \\
        peptic\_ulcer                    & Negative                               & 99.97\% [39805]           & 100.00\% [3188]           & 99.97\% [42993]     \\
                                         & Positive                               & 0.03\% [12]               & 0.00\% [0]                & 0.03\% [12]         \\
        aids                             & Negative                               & 99.90\% [39776]           & 99.84\% [3183]            & 99.89\% [42959]     \\
                                         & Positive                               & 0.10\% [41]               & 0.16\% [5]                & 0.11\% [46]         \\
        lymphoma                         & Negative                               & 99.80\% [39739]           & 99.40\% [3169]            & 99.77\% [42908]     \\
                                         & Positive                               & 0.20\% [78]               & 0.60\% [19]               & 0.23\% [97]         \\
        metastatic\_cancer               & Negative                               & 99.09\% [39456]           & 98.84\% [3151]            & 99.07\% [42607]     \\
                                         & Positive                               & 0.91\% [361]              & 1.16\% [37]               & 0.93\% [398]        \\
        solid\_tumor                     & Negative                               & 97.80\% [38943]           & 97.24\% [3100]            & 97.76\% [42043]     \\
                                         & Positive                               & 2.20\% [874]              & 2.76\% [88]               & 2.24\% [962]        \\
        rheumatoid\_arthritis            & Negative                               & 99.97\% [39805]           & 99.87\% [3184]            & 99.96\% [42989]     \\
                                         & Positive                               & 0.03\% [12]               & 0.13\% [4]                & 0.04\% [16]         \\
        coagulopathy                     & Negative                               & 99.96\% [39801]           & 99.97\% [3187]            & 99.96\% [42988]     \\
                                         & Positive                               & 0.04\% [16]               & 0.03\% [1]                & 0.04\% [17]         \\
        obesity                          & Negative                               & 99.94\% [39792]           & 100.00\% [3188]           & 99.94\% [42980]     \\
                                         & Positive                               & 0.06\% [25]               & 0.00\% [0]                & 0.06\% [25]         \\
        weight\_loss                     & Negative                               & 99.99\% [39813]           & 100.00\% [3188]           & 99.99\% [43001]     \\
                                         & Positive                               & 0.01\% [4]                & 0.00\% [0]                & 0.01\% [4]          \\
        fluid\_electrolyte               & Negative                               & 99.66\% [39682]           & 99.87\% [3184]            & 99.68\% [42866]     \\
                                         & Positive                               & 0.34\% [135]              & 0.13\% [4]                & 0.32\% [139]        \\
        blood\_loss\_anemia              & Negative                               & 99.97\% [39807]           & 99.97\% [3187]            & 99.97\% [42994]     \\
                                         & Positive                               & 0.03\% [10]               & 0.03\% [1]                & 0.03\% [11]         \\
        deficiency\_anemias              & Negative                               & 99.98\% [39808]           & 100.00\% [3188]           & 99.98\% [42996]     \\
                                         & Positive                               & 0.02\% [9]                & 0.00\% [0]                & 0.02\% [9]          \\
        alcohol\_abuse                   & Negative                               & 99.54\% [39635]           & 100.00\% [3188]           & 99.58\% [42823]     \\
                                         & Positive                               & 0.46\% [182]              & 0.00\% [0]                & 0.42\% [182]        \\
        drug\_abuse                      & Negative                               & 99.94\% [39794]           & 100.00\% [3188]           & 99.95\% [42982]     \\
                                         & Positive                               & 0.06\% [23]               & 0.00\% [0]                & 0.05\% [23]         \\
        psychoses                        & Negative                               & 99.97\% [39807]           & 100.00\% [3188]           & 99.98\% [42995]     \\
                                         & Positive                               & 0.03\% [10]               & 0.00\% [0]                & 0.02\% [10]         \\
        depression                       & Negative                               & 100.00\% [39817]          & 100.00\% [3188]           & 100.00\% [43005]    \\
                                         & Positive                               & 0.00\% [0]                & 0.00\% [0]                & 0.00\% [0]          \\
        \bottomrule
    \end{longtable}}

\end{document}